\documentclass[12pt]{article}

\usepackage{PRIMEarxiv}
\usepackage[utf8]{inputenc}
\usepackage[T1]{fontenc}
\usepackage{hyperref}
\hypersetup{hidelinks}
\usepackage{url}
\usepackage[noadjust]{cite}
\usepackage{booktabs}
\usepackage{tabularx}
\usepackage{array}
\usepackage{amsfonts}
\usepackage{amsmath}
\usepackage{nicefrac}
\usepackage{microtype}
\usepackage{fancyhdr}
\usepackage{graphicx}
\usepackage{xcolor}
\usepackage{subcaption}
\usepackage{float}

\usepackage{CJKutf8}
\usepackage{tikz}
\usepackage{pgfplots}
\pgfplotsset{compat=1.18}
\usepgfplotslibrary{groupplots}
\usetikzlibrary{shapes.geometric, arrows.meta, positioning, fit, backgrounds, calc}

\definecolor{anaC}{RGB}{31,119,180}
\definecolor{evoC}{RGB}{44,160,44}
\definecolor{memC}{RGB}{214,39,40}
\definecolor{cmpC}{RGB}{148,103,189}

\tikzset{
  proc/.style   = {rectangle, rounded corners=2pt, draw=black!70, fill=blue!6,
                   text width=3.4cm, align=center, minimum height=0.9cm, font=\small},
  dec/.style    = {diamond, aspect=2, draw=black!70, fill=orange!12,
                   text width=2.4cm, align=center, inner sep=1pt, font=\small},
  term/.style   = {rectangle, draw=black!70, fill=green!8, text width=3.6cm,
                   align=center, minimum height=0.9cm, font=\small},
  phase/.style  = {rectangle, rounded corners=2pt, draw=black!70, fill=blue!8,
                   text width=10.5cm, align=left, minimum height=0.8cm, font=\small},
  flow/.style   = {-{Stealth[length=2mm]}, draw=black!60, thick},
  box/.style   = {rounded corners=2pt, draw=black!65, align=center, font=\small,
                  minimum height=0.85cm, inner sep=4pt},
  ana/.style   = {box, fill=anaC!10, draw=anaC},
  evo/.style   = {box, fill=evoC!10, draw=evoC},
  dec/.style   = {diamond, aspect=2.1, draw=black!65, fill=orange!12, align=center,
                  font=\scriptsize, inner sep=1pt},
  fl/.style    = {-{Stealth[length=2mm]}, thick, draw=black!60},
}

\graphicspath{{media/}{figures/}}

\title{CANNBot-CAKE: Collaborative Agentic Kernel Evolution for Ascend NPU}
\author{%
\parbox{0.95\textwidth}{\centering
Deng~Pan\textsuperscript{*}, Xue-Jian~Gao\textsuperscript{*,$\dagger$},
Junyi~Fan, Bin~Du, Jiasheng~Li, Xinxian~Chen,
Nok~Yan~Chang, Xinlin~Li, 
Lingchao~Zheng, Jun~Li,
Yuwei~Fan\textsuperscript{$\ddagger$}\\[0.4ex]
\normalsize\normalfont Huawei
}%
}

\newcommand{\symfootnote}[2]{%
  \begingroup
  \renewcommand{\thefootnote}{#1}%
  \begin{NoHyper}%
  \footnotetext[0]{#2}%
  \end{NoHyper}%
  \endgroup
}

\begin{document}
\begin{CJK*}{UTF8}{gbsn}
\maketitle
\symfootnote{*}{These authors contributed equally: Deng Pan, Xue-Jian Gao.}
\symfootnote{$\dagger$}{\texttt{gao.xuejian1@huawei.com}}
\symfootnote{$\ddagger$}{\texttt{fanyuwei2@huawei.com}}
\setlength{\headheight}{14.5pt}
\setcounter{footnote}{0}

\begin{abstract}
Efficient custom operators are essential for modern deep learning workloads, yet
developing high-performance kernels for Ascend NPUs remains slow,
expert-intensive, and difficult to automate. Although AI agents now approach expert performance on GPU kernel benchmarks, current models remain less capable of generating AscendC kernels for NPUs, potentially reflecting the relative scarcity of relevant training data. We thus present CAKE, a CANN-native system for deployable custom operators
that combines adaptive multi-DSL routing, separate Generation and Evolution
modes, and reviewed knowledge reuse through CAKE Book and a Self-Evolving Agent.
On 53 CANN-Bench v1.1 operators on Ascend 950PR, CAKE's aggregated results
achieve $100\%$ case pass rate on the standard set (1,060/1,060) and $98.28\%$
on the hidden set (4,167/4,240), with an overall mean score of 78.63 out of 100 and an average performance speedup of $1.334\times$ over the baseline.
Evolution then pushes correct operators further:
EngramGateFusion cuts aggregate device time by $46.6\%$ over 20 rounds and
roughly halves the mean roofline gap. These results offer a practical path
toward closing Ascend's AI-for-kernel capability gap.
\end{abstract}

\keywords{Agentic systems \and Ascend NPU \and CANN \and AscendC \and Kernel generation \and Evolutionary optimization}

\section{Motivation and System Overview}
\label{sec:introduction}

An accelerator reaches its advertised throughput only when every operator in a
model is backed by an efficient, hardware-tuned compute kernel. Kernel quality,
not a chip's peak throughput, sets end-to-end performance, and writing these
kernels by hand is slow and demands scarce low-level expertise. Automating that
work has become an active research target.

Large language models have made that target newly reachable: coupled with
execution feedback and search, they can generate correct and competitive GPU
kernels. This progress rests on a favorable ecosystem. CUDA and Triton expose
abundant open kernels, documented optimization idioms, and large public code
corpora, so a model can recognize and adapt patterns it has effectively already
seen. NPUs such as Ascend offer no comparable foundation; their vendor
programming model, AscendC, appears only sparsely in public data. In published
evaluations, general-purpose models succeed on
essentially \emph{none} of the complex AscendC kernels they attempt in a
zero-shot setting~\cite{cao2026ascendkernelgen}, and one-shot AscendC pass rates
remain at or below $2.1\%$ where CUDA reaches
$44.2$--$52.6\%$~\cite{multikernelbench2025,ascendoptimizer2026}. These results
motivate supplementing pretrained models with Ascend-specific programming
knowledge and execution feedback.

Ascend kernel development requires coordinating heterogeneous compute units
over an explicitly managed on-chip memory hierarchy, with data movement, pipeline
overlap, and synchronization all under program control
(Appendix~\ref{app:a5} describes the additional programming paths on Ascend 950). The
consequence for a code-generating agent is that an Ascend operator is not a single
kernel but a \emph{two-part artifact}~\cite{ascendoptimizer2026}: a host-side
tiling program that decides how data are partitioned and moved, and a device-side
kernel that decides how computation is scheduled. Performance is a property of the
coordination between these two components: tiling determines the data available
to each core, while the device kernel schedules computation and transfers.

In practice the bar is higher still. The unit an engineer can actually ship is
neither a host program nor a device kernel in isolation, but a \emph{deployable
CANN custom operator}: one that builds, packages, installs, binds to Python,
passes precision checks against a reference, and profiles on real hardware.
Recent Ascend systems each automate one slice of this chain
(Section~\ref{sec:related} and Table~\ref{tab:closest-work})---training a model,
transcompiling a domain-specific language (DSL), optimizing an existing kernel, or supplying curated domain
knowledge to an evolutionary search. The last of these, AgenticCANN, is the
closest prior system to ours and the one that most directly confronts the corpus
problem: it organizes AscendC expertise into a six-level taxonomy, from hardware
mental model down to verified kernel exemplars, and injects the levels selectively
by evolutionary phase, raising the share of elementwise kernels it can synthesize
from $57\%$ to $86\%$ on Ascend 910B~\cite{agenticcann2026}. This demonstrates
the value of explicit domain knowledge. Its reported feasibility also varies
by operator family: $90$--$100\%$ on elementwise and normalization operators,
but $56\%$ on fusion, over a private $54$-operator sample.

These results motivate a workflow that carries generated code through packaging
and deployment while reusing the results of previous runs. Such a workflow needs
to retain intermediate artifacts for debugging, use evaluation outcomes to guide
generation, and make useful lessons available to later tasks.

We present \textbf{CANNBot-CAKE} (hereafter CAKE), an open-source CANN-native
agentic workflow released in CANNBot, Huawei's official repository for Ascend
agents.\footnote{\url{https://gitcode.com/cann/cannbot-skills}} CAKE generates
and evolves deployable Ascend operators from either a natural-language
requirement or an existing AscendC baseline. CAKE uses existing language models
to coordinate these stages. A request becomes an
operator contract, then a PyTorch reference, a functional form, a routed DSL
baseline, host-side tiling, device AscendC code, a CANN package, a Python
binding, and an evaluation record---each a durable object that can be inspected,
validated, and reused. A multi-DSL \emph{router} chooses among four
code-generation tracks using the operator's structural features and previous
evaluation scores. It also explores less-tested tracks that may perform better.
\textbf{Generation mode} produces one deployable operator through generation,
validation, and repair. A profiling assistant, the \emph{Advisor}, diagnoses
performance bottlenecks and suggests changes.
\textbf{Evolution mode} (CAKE-evo) runs a multi-round campaign that reuses shared artifacts,
spawns isolated variants, and ranks them by compilation, precision, and
performance. Candidates are scored by automated evaluation on
CAKE-Bench, a user-defined evaluation suite that we instantiate with the official
CANN-Bench task set~\cite{cannbench}; a \emph{self-evolving agent} mines
those measured runs to improve the router, skills, and repair templates; and a
cross-user knowledge base, CAKE Book, accumulates verified lessons in
interlinked wiki pages~\cite{ming2026llmwiki}.

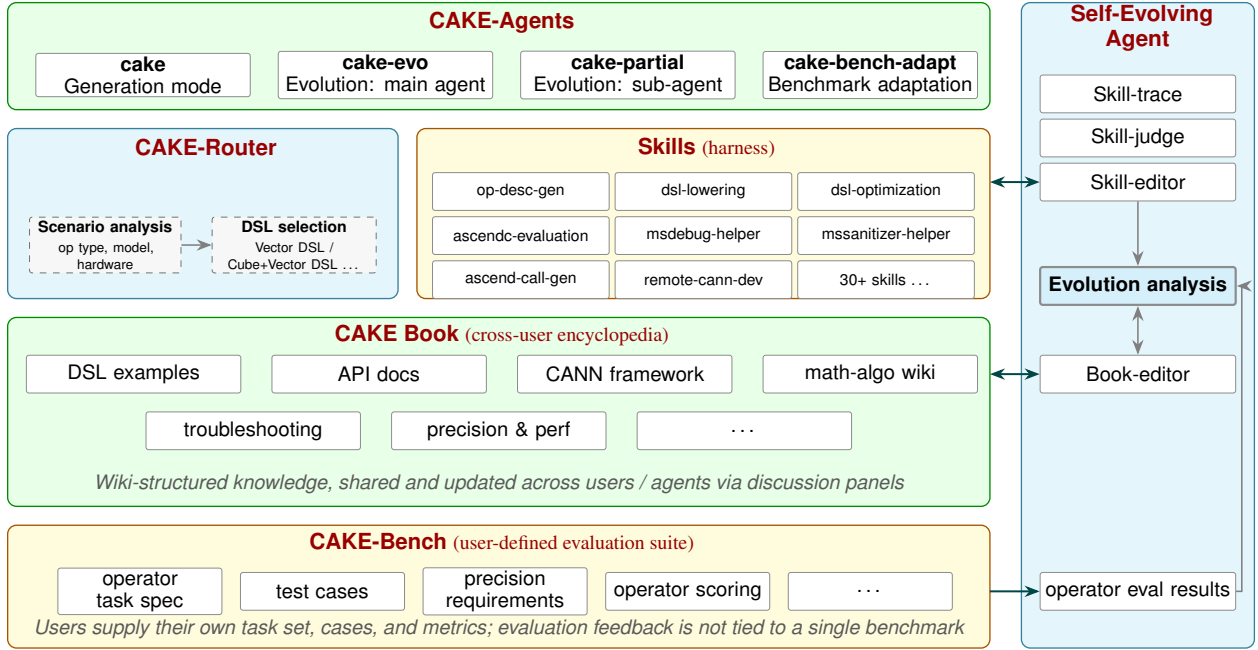
\begin{figure}[t]
  \centering
  \resizebox{\linewidth}{!}{%
  \begin{tikzpicture}[
    font=\sffamily,
    >={Stealth[width=1.8mm,length=2.4mm]},
    cbox/.style={draw=black!45, fill=white, rounded corners=1.2pt, align=center,
                 font=\sffamily\scriptsize, inner sep=1.5pt, minimum height=6mm},
    dbox/.style={cbox, dashed, draw=black!55, fill=black!3},
    ttl/.style={font=\sffamily\bfseries\small, text=red!60!black},
    tnorm/.style={font=\sffamily\scriptsize, text=red!60!black},
    sub/.style={font=\sffamily\itshape\scriptsize, text=black!65, align=center},
    sm/.style={font=\sffamily\fontsize{7pt}{8pt}\selectfont},
    bandG/.style={rounded corners=4pt, draw=green!55!black, fill=green!9, line width=0.6pt},
    bandB/.style={rounded corners=4pt, draw=cyan!55!black, fill=cyan!9, line width=0.6pt},
    bandY/.style={rounded corners=4pt, draw=orange!65!black, fill=yellow!16, line width=0.6pt},
    conn/.style={<->, line width=0.9pt, teal!55!black},
    flow/.style={->, line width=0.7pt, black!50},
  ]
    \draw[bandG] (0,5.7) rectangle (15.6,7.4);
    \node[ttl] at (7.8,7.08) {CAKE-Agents};
    \foreach \x/\nm/\d in {%
        2.15/cake/{Generation mode},
        6.0/cake-evo/{Evolution: main agent},
        9.85/cake-partial/{Evolution: sub-agent},
        13.7/cake-bench-adapt/{Benchmark adaptation}} {
      \node[cbox, text width=33mm] at (\x,6.25) {\textbf{\nm}\\\d};
    }

    \draw[bandB] (0,2.7) rectangle (6.2,5.4);
    \node[ttl] at (3.1,5.1) {CAKE-Router};
    \node[dbox, sm, text width=23mm] (scen) at (1.55,3.55)
      {\textbf{Scenario analysis}\\[2pt]{\fontsize{6pt}{7pt}\selectfont op type, model,\\ hardware}};
    \node[dbox, sm, text width=25mm] (sel) at (4.55,3.55)
      {\textbf{DSL selection}\\[2pt]{\fontsize{6pt}{7pt}\selectfont Vector DSL /\\ Cube+Vector DSL \ldots}};
    \draw[flow] (scen) -- (sel);

    \draw[bandY] (6.5,2.7) rectangle (15.6,5.4);
    \node[ttl] at (11.05,5.1) {Skills {\normalfont\scriptsize (harness)}};
    \foreach \x/\y/\name in {%
        8.15/4.4/op-desc-gen,      11.05/4.4/dsl-lowering,    13.95/4.4/dsl-optimization,
        8.15/3.7/ascendc-evaluation,11.05/3.7/msdebug-helper, 13.95/3.7/mssanitizer-helper,
        8.15/3.0/ascend-call-gen,  11.05/3.0/remote-cann-dev, 13.95/3.0/{30+ skills \ldots}} {
      \node[cbox, sm, text width=27mm] at (\x,\y) {\name};
    }

    \draw[bandG] (0,-0.6) rectangle (15.6,2.4);
    \node[ttl] at (7.8,2.12) {CAKE Book {\normalfont\scriptsize (cross-user encyclopedia)}};
    \foreach \x/\name in {%
        2.0/DSL examples, 5.9/API docs, 9.8/CANN framework, 13.7/{math-algo wiki}} {
      \node[cbox, text width=33mm] at (\x,1.5) {\name};
    }
    \foreach \x/\name in {%
        3.9/troubleshooting, 7.8/{precision \& perf}, 11.7/\ldots} {
      \node[cbox, text width=33mm] at (\x,0.6) {\name};
    }
    \node[sub] at (7.8,-0.25)
      {Wiki-structured knowledge, shared and updated across users / agents via discussion panels};

    \draw[bandY] (0,-2.85) rectangle (15.6,-0.9);
    \node[ttl] at (7.8,-1.18) {CAKE-Bench {\normalfont\scriptsize (user-defined evaluation suite)}};
    \foreach \x/\name in {%
        2.1/{operator task spec}, 5.0/{test cases}, 7.9/{precision requirements},
        10.8/{operator scoring}, 13.7/\ldots} {
      \node[cbox, text width=25mm] at (\x,-1.95) {\name};
    }
    \node[sub] at (7.8,-2.55)
      {Users supply their own task set, cases, and metrics; evaluation feedback is not tied to a single benchmark};

    \draw[bandB] (16.1,-2.85) rectangle (19.8,7.4);
    \node[ttl, text width=32mm, align=center] at (17.95,7.0) {Self-Evolving Agent};
    \node[cbox, text width=30mm] (strace) at (17.95,5.95) {Skill-trace};
    \node[cbox, text width=30mm] (sjudge) at (17.95,5.25) {Skill-judge};
    \node[cbox, text width=30mm] (sedit)  at (17.95,4.55) {Skill-editor};
    \node[cbox, text width=30mm, fill=cyan!14, line width=1pt] (evol) at (17.95,2.9)
      {\textbf{Evolution analysis}};
    \node[cbox, text width=30mm] (bedit)  at (17.95,1.5) {Book-editor};
    \node[cbox, text width=30mm] (eres)   at (17.95,-1.95) {operator eval results};
    \draw[flow] (sedit) -- (evol);
    \draw[<->, line width=0.7pt, black!50] (evol) -- (bedit);
    \draw[flow] (eres.east) -- (19.6,-1.95) -- (19.6,2.9) -- (evol.east); %

    \draw[conn] (15.6,4.55) -- (sedit.west);   %
    \draw[conn] (15.6,1.5)  -- (bedit.west);    %
    \draw[flow, line width=0.9pt, teal!55!black] (15.6,-1.95) -- (eres.west); %
  \end{tikzpicture}}
  \caption{CAKE architecture. Four stacked layers cooperate to turn an operator
  request into a deployable, benchmark-validated Ascend operator.
  \textbf{CAKE-Agents} provides the controllers---generation, the
  evolution main and sub-agents, and automatic benchmark adaptation. The
  \textbf{CAKE-Router} gates each request on track--chip--operator-type
  feasibility, then scores the surviving code-generation tracks against the
  operator's eight-dimensional structural profile, while the \textbf{Skills}
  harness supplies typed lifecycle tools for description, lowering, optimization,
  evaluation, and debugging. The \textbf{CAKE Book} is a cross-user encyclopedia
  maintained as a shared wiki, and \textbf{CAKE-Bench} is a user-defined
  evaluation suite---operator task specifications, test cases, precision
  requirements, and scoring---so that evaluation feedback is not tied to any
  single benchmark. A \textbf{Self-Evolving Agent} observes all layers: it
  traces, judges, and edits skills, analyses evolution runs, writes verified
  experience back to CAKE Book, and closes the loop with measured operator
  results.}
  \label{fig:architecture}
\end{figure}

Figure~\ref{fig:architecture} summarizes five components of CAKE.
First, a CANN-native staged pipeline (CAKE-Agents) turns operator
intent into a deployable custom operator through Generation and Evolution modes,
exposing every stage---contract, reference, DSL, host tiling, kernel, package,
binding, and evaluation---as an inspectable object. Second, the CAKE-Router makes
DSL choice an explicit, measured decision over four code-generation tracks: a
capability gate removes infeasible track--chip--operator-type combinations, a
per-track item-response model turns past evaluated operators into ability
estimates adjusted for operator difficulty, and an
information-gated exploration rule allocates additional evaluations to
tracks that could overtake the incumbent (Section~\ref{subsec:dsl-routing}). Third, a Skills harness of more than thirty
typed lifecycle tools---for description, lowering, optimization, evaluation, and
debugging---gives the agents a reusable substrate for operator engineering
(Section~\ref{subsec:skills-harness}).
Fourth, CAKE Book accumulates verified CANN experience as a cross-user wiki,
shared and updated across users and agents and queried before generation. Fifth,
a self-evolving agent closes the loop:
continuous automated evaluation on CAKE-Bench produces measured signals that an
analysis module mines to trace, judge, and edit skills and to promote
evidence-gated improvements back into the router, prompts, repair templates, and
CAKE Book. The evidence in this report combines aggregated standard- and
hidden-set results on 53 CANN-Bench operators on Ascend 950PR, replayable
Router decisions, and operational
evidence from the CAKE Book write path.
The EngramGateFusion case study demonstrates multi-round optimization of a
generated kernel. Section~\ref{sec:scope-roadmap} outlines further development
of the workflow and its evaluation.

Table~\ref{tab:closest-work} contrasts CAKE with the closest Ascend-native
systems by their generation, optimization, delivery, and knowledge-reuse mechanisms.

\begin{table}[t]
  \centering
  \footnotesize
  \setlength{\tabcolsep}{4pt}
  \renewcommand{\arraystretch}{1.05}
  \newcolumntype{Y}{>{\raggedright\arraybackslash}X}
  \newcolumntype{Q}{>{\hsize=1.10\hsize\raggedright\arraybackslash}X}
  \newcolumntype{N}{>{\hsize=0.85\hsize\raggedright\arraybackslash}X}
  \newcolumntype{M}{>{\hsize=0.95\hsize\raggedright\arraybackslash}X}
  \begin{tabularx}{\linewidth}{@{}l Q N M Q@{}}
    \toprule
    \textbf{System} & \textbf{Core approach} & \textbf{Input} &
    \textbf{Delivered unit} & \textbf{Search \& cross-run knowledge} \\
    \midrule
    AscendKernelGen~\cite{cao2026ascendkernelgen} &
    Train a domain model (Ascend-CoT data, SFT/RL) &
    NL requirement &
    AscendC kernel scored on a benchmark &
    None; experience fixed in model weights \\
    \addlinespace[2pt]
    AscendCraft~\cite{ascendcraft2026} &
    DSL generation then constrained lowering passes &
    NL requirement &
    AscendC kernel &
    None; reuse via category expert examples \\
    \addlinespace[2pt]
    AscendOptimizer~\cite{ascendoptimizer2026} &
    Episodic agent: optimization rewind + tiling search &
    Existing correct AscendC operator &
    Optimized kernel and host tiling &
    Host-tiling evolution; per-benchmark experience bank \\
    \addlinespace[2pt]
    AgenticCANN~\cite{agenticcann2026} &
    Six-level expert knowledge taxonomy injected by phase; stage-adaptive agent scheduling &
    NL requirement &
    Compiled and numerically verified AscendC kernel &
    Candidate evolution; knowledge authored once, not written back from runs \\
    \addlinespace[2pt]
    \textbf{CAKE (ours)} &
    Agentic workflow over staged, inspectable artifacts &
    NL requirement \emph{or} AscendC baseline &
    Deployable CANN operator (build, install, bind, evaluate) &
    Constrained variant evolution with Good/Medium/Poor sampling; CAKE Book wiki across runs and agents \\
    \bottomrule
  \end{tabularx}
  \caption{Closest Ascend-native operator systems versus CAKE. Each prior system
  targets one slice of operator engineering---training a model, transcompiling a
  DSL, optimizing an existing kernel, or supplying curated knowledge to a
  search---whereas CAKE delivers a deployable CANN operator and reuses measured
  experience across runs. AgenticCANN is the closest: it also treats the scarce
  AscendC corpus as the central obstacle, but its knowledge is authored by experts
  ahead of time rather than accumulated from measured runs. Entries are drawn from
  the cited papers.}
  \label{tab:closest-work}
\end{table}

\subsection{Related Systems}
\label{sec:related}

\paragraph{Kernel benchmarks and feedback-driven kernel agents.}
Benchmarks such as KernelBench, TritonBench, and MultiKernelBench reframed kernel
synthesis around measured compile, correctness, and performance
funnels~\cite{ouyang2025kernelbench,li2025tritonbench,multikernelbench2025}, and a
wave of GPU kernel agents---among them KernelAgent, CudaForge, AutoKernel,
KernelFoundry, and KernelEvolve---showed that hardware feedback and search drive
synthesis past one-shot
prompting~\cite{kernelagent2026,cudaforge2025,autokernel2026,kernelfoundry2026,kernelevolve2025}.
On Ascend, related systems use several approaches to supply domain expertise.
AscendKernelGen trains a
domain-adapted model on a reasoning dataset to emit AscendC
code~\cite{cao2026ascendkernelgen}; AscendCraft transcompiles a lightweight DSL
into AscendC through constrained lowering passes~\cite{ascendcraft2026};
AscendOptimizer improves an already correct AscendC implementation through
rewind-distilled experience and tiling search~\cite{ascendoptimizer2026}; and
AgenticCANN attacks the corpus scarcity head-on, encoding AscendC expertise as a
six-level taxonomy---hardware model, math-to-code mapping, execution invariants,
tiling formulas, API signatures, verified exemplars---and admitting different
levels at different phases of an evolutionary search, while scheduling
high-exploration agents for seed discovery and cheaper agents for
refinement~\cite{agenticcann2026}. Its ablations measure the contribution of
this knowledge: adding
knowledge levels lifts elementwise feasibility from $57\%$ to $86\%$, with
hardware-constraint knowledge alone worth $20$ percentage points on LayerNorm. The
difference relevant to CAKE is how this knowledge is maintained. AgenticCANN
uses an expert-curated taxonomy; CAKE Book also collects entries
from measured runs and promotes them only after cross-user review and on-device
verification, which is what allows later operators to benefit from earlier
failures. Table~\ref{tab:closest-work} compares these systems' delivery scope;
CAKE targets the chain from intent to a
deployable, benchmark-validated operator.

\paragraph{Routing.}
Learned routers such as RouteLLM dispatch a query to a stronger or weaker
model~\cite{routellm}, and compiler auto-schedulers such as Ansor use learned
cost models to select a schedule within a single intermediate
representation~\cite{ansor}. Kernel-generation agents, by contrast, are largely
single-DSL~\cite{ascendcraft2026,cao2026ascendkernelgen,cudaforge2025}; the few
that span several backends either expose support for multiple DSLs without a
published selection rule or defer the choice to an opaque per-kernel agent
decision~\cite{du2025akgkernelagent,autokernel2026}. CAKE instead scores four peer
code-generation tracks under a published rule: an item-response model estimates
each track's ability per structural dimension from past evaluated operators, so
that the ordering of tracks changes with the operator rather than being fixed,
and every decision is recorded with the candidate estimates that produced it
(Section~\ref{subsec:dsl-routing}).

\paragraph{Agent memory and knowledge bases.}
Agent memory has progressed from per-episode self-reflection~\cite{reflexion} and
single-agent skill or experience libraries~\cite{voyager,expel} to self-linking
memory notes~\cite{amem}, and recently to sharing across actors through banked
lessons for cooperating code agents~\cite{lessonslearned} and provenance-tracked
multi-user memory~\cite{collaborativememory}. In the same Ascend domain,
AscendOptimizer mines a single-agent experience bank for kernel
rewriting~\cite{ascendoptimizer2026}, and AgenticCANN supplies a hand-authored
taxonomy of AscendC knowledge as retrieval context~\cite{agenticcann2026}.
CAKE Book combines three properties: it is an
agent-maintained wiki of interlinked pages with an explicit error notebook rather
than a flat store; it is written by the runs it serves, so its contents grow with
measured evidence rather than with expert authoring effort; and its entries become
canonical only after cross-user, human review and on-device verification.

\paragraph{Self-evolving agents.}
A growing body of work studies agents that improve themselves: recursive
self-referential systems rewrite their own scaffolding~\cite{stop,godelagent},
automated agent-design methods search the space of agentic
architectures~\cite{adas}, and evolutionary approaches maintain populations of
self-modifying agents validated by benchmark fitness~\cite{darwingodel}. In the
kernel domain, agentic systems evolve archives of optimized
kernels~\cite{aicudaengineer}. CAKE's Self-Evolving Agent targets the generation
workflow. It mines
many completed generation and profiling runs to promote evidence-gated
improvements into the router, skills, repair templates, and CAKE Book.

\section{From Intent to Deployable Operator}
\label{sec:generation-pipeline}

CAKE divides operator development into stages with explicit inputs, outputs,
and validation checks. Each stage produces an artifact used by subsequent
stages, allowing failures to be diagnosed before the workflow proceeds.

\begin{itemize}
  \item \textbf{Operator contract generation.} The request (or an \texttt{api\_description.md}) is compiled into an \emph{operator contract} (\texttt{op\_desc.json}): the single source of truth for the operator's identity, semantic formulas, representative I/O with dtype/shape constraints, attributes, and case table. Every later stage derives its signature from it, including the schema features that are otherwise lost downstream --- \texttt{is\_list}, \texttt{optional}, string-attribute \texttt{enum}s, \texttt{output\_shape} derivations, and channel-first \texttt{layout}.

  \item \textbf{Reference and functional forms.} CAKE builds a PyTorch reference, preserving a supplied implementation verbatim and fixing random seeds for reproducibility. It then converts the reference to a stateless \texttt{module\_fn}. The reference defines expected outputs for correctness checks, the test inputs exercise both reference and generated code, and the functional form exposes dataflow for routing and lowering.

  \item \textbf{Multi-DSL router.} The router represents the contract and functional form as an eight-dimensional structural profile, removes tracks that cannot support the operator type or target device, and scores the remaining tracks using previous evaluations. Section~\ref{subsec:dsl-routing} describes the model. Users can instead specify a track directly through the contract's \texttt{dsl\_track} field.

  \item \textbf{Lowering and project construction.} The selected track lowers the functional form into host-side tiling and device-side code while preserving the contract as the interface authority. CAKE materializes the generated source, build configuration, tiling metadata, wrappers, and project skeleton as separate artifacts so that a failure can be localized to a stage rather than attributed to an opaque model response.

  \item \textbf{Package, install, and bind.} A candidate is built as a CANN custom-operator package, installed into the pinned environment, and exposed through the generated Python binding. These stages are part of the deliverable rather than post-processing: code that compiles as a kernel snippet but cannot be packaged or invoked does not count as a completed operator.

  \item \textbf{Evaluate and repair.} The installed operator is executed against the reference under dtype-aware precision checks and on-device timing. Compilation, packaging, binding, correctness, and profiling results are written next to the candidate. The resulting evidence routes the next action: build failures return to lowering or project construction, numerical failures return to the contract, reference, or tiling assumptions, and correct but slow candidates enter profiling-guided repair or Evolution mode.
  \end{itemize}

The unit of progress in this lifecycle is therefore an artifact plus its validation
record. A run can stop at a failed intermediate stage without losing the evidence
needed to diagnose it, while a successful run leaves a package, binding, test
record, profile, and provenance trail that another engineer or agent can inspect.
These records support both optimization within the current run and knowledge
reuse across runs.

\subsection{The Skills harness}
\label{subsec:skills-harness}

The stages above describe what is produced; the Skills harness is what executes
them. A skill is a typed lifecycle tool: it names the artifact it consumes, the
artifact it writes, and the condition under which its output is admissible. The
orchestrating agent does not contain the procedures --- it declares the skills it
may call, currently more than thirty, and the runtime resolves each call against
that declaration. Adding a capability therefore means adding a skill and listing
it, not editing the agent.

The harness coordinates skill selection, validation, and failure recovery.

\paragraph{Stage binding, with track-specific variants.}
Skills are bound to stages rather than invoked ad hoc. Contract, reference,
functional conversion, and tiling pre-classification are shared by every run; the
router then selects a code-generation track, and the lowering stages dispatch to
the variant of each skill specialized for that track --- separate lowering and
call-generation skills exist for the Vector and the fused Cube--Vector paths, so
that the difference between them lives in the harness rather than inside a
branching prompt. Review, evaluation, Advisor-guided repair, and run retrospective
are shared again. A track is thus a path through a fixed skill graph, which is why
the router's choice is auditable after the fact.

\paragraph{Stage validation.}
A stage advances only when its artifact passes a machine-checked gate. The
contract illustrates these checks: before any code is generated, the
\texttt{op\_desc} is checked for dtype coverage against the full case table, for
signature agreement with the task's declared schema, and for the shape metadata
that packaging later depends on. These checks catch incomplete dtype coverage,
argument mismatches, and missing output-shape metadata before they propagate
to compilation or packaging. A failed check returns control to the stage that
produced the artifact.

\paragraph{Execution tracing and failure recovery.}
Two cross-cutting skills record execution and coordinate repairs throughout a run.
Every skill call is bracketed by a trace record, so a run leaves a complete,
machine-readable account of which skills fired, in what order, and with what
outcome --- this trace is the input the self-evolving agent of
Section~\ref{sec:self-evolving-agent} reads to identify recurring skill failures.
Alongside it, a
test-driven meta-skill governs what happens on failure: a diagnosis is treated as
a hypothesis that stays open until an end-to-end pass closes it, and a failure
mode seen twice must be pushed back into the lowering step or the project
skeleton that permitted it rather than accumulated as another entry in a pitfall
list. Retries are bounded, and a stage that does not pass its gate produces no
commit, so the version history records the stages that cleared their checks.

\section{Adaptive Routing and Generation}
\label{subsec:dsl-routing}

CAKE carries four code-generation \emph{tracks} that can each produce a
deployable operator: \textbf{DSL-A}, the mature AscendC-lowering track that
covers both vector and cube--vector operators on the 910 series;
\textbf{DSL-B}, a high-level tile DSL whose strength is cube--vector topology;
\textbf{DSL-C}, the vector and cube--vector generation track for the 950; and the
\textbf{SIMT} track detailed in Appendix~\ref{app:a5-simt}. A track is an
\emph{implementation route}, not an operator class: whether an operator is
vector, cube--vector, or SIMT-shaped is an input to the routing decision, and
several tracks compete for the same class.

\begin{figure}[t]
\centering
\resizebox{\linewidth}{!}{%
\begin{tikzpicture}[
    sproc/.style={rectangle, rounded corners=2pt, draw=black!70, fill=blue!6,
                  align=center, minimum height=0.85cm, font=\small, inner sep=3pt},
    sdec/.style={diamond, aspect=2.1, draw=black!70, fill=orange!12,
                 align=center, inner sep=0pt, font=\small},
    sterm/.style={rectangle, draw=black!70, fill=green!8, align=center,
                  minimum height=0.85cm, font=\small, inner sep=3pt},
    lay/.style={rectangle, rounded corners=2pt, draw=black!45, fill=black!5,
                minimum width=0.85cm, minimum height=0.6cm, font=\small\bfseries},
    ar/.style={-{Stealth[length=2mm]}, draw=black!60, thick}]

  \node[sproc, text width=4.8cm] (in)   at (0,0)      {\texttt{op\_desc.json} $+$ functional form $+$ device};
  \node[sproc, text width=4.8cm] (feat) at (0,-1.5)   {$\varphi \in \mathbb{Z}^{8}$, operator type, $b = \beta^{\top}\varphi$~~(Eq.~\ref{eq:difficulty})};
  \node[sdec,  text width=1.9cm] (gate) at (0,-3.1)   {feasible\\tracks?};
  \node[sterm, text width=3.3cm, fill=orange!10] (defer) at (-4.3,-4.8) {\textbf{abstain}\\track menu $\rightarrow$ model};
  \node[sproc, text width=4.0cm] (score) at (2.6,-4.8) {per-track $p_{\text{track}}(\varphi)$~~(Eq.~\ref{eq:mirt})};
  \node[sdec,  text width=1.9cm] (voi)   at (2.6,-6.5) {contender\\eligible?};
  \node[sterm, text width=2.8cm] (exploit) at (0.1,-8.2) {\textbf{exploit}\\$\arg\max p$};
  \node[sterm, text width=3.4cm, fill=green!16] (expl) at (5.2,-8.2) {\textbf{explore}\\$\arg\max$ upper bound};
  \node[sproc, text width=6.6cm, fill=blue!11] (ledger) at (2.6,-9.9) {ledger row: $(\text{track},\ \text{chip},\ \varphi,\ b,\ y,\ \text{skill version})$};

  \draw[ar] (in)   -- (feat);
  \draw[ar] (feat) -- (gate);
  \draw[ar] (gate) -| (defer) node[pos=0.30, above, font=\small] {none};
  \draw[ar] (gate) -| (score) node[pos=0.30, above, font=\small] {$\geq 1$};
  \draw[ar] (score) -- (voi);
  \draw[ar] (voi) -| (exploit) node[pos=0.30, above, font=\small] {no};
  \draw[ar] (voi) -| (expl)    node[pos=0.30, above, font=\small] {yes};
  \draw[ar] (exploit.south) -- (exploit.south |- ledger.north);
  \draw[ar] (expl.south)    -- (expl.south |- ledger.north);
  \draw[ar] (ledger.east) -- (10.2,-9.9) -- (10.2,-4.8) -- (score.east)
        node[pos=0.45, right, font=\small, align=left]
        {evaluated score $y$\\updates $\theta$};

  \node[lay] at (-6.7,-1.5) {L1};
  \node[lay] at (-6.7,-3.1) {L0};
  \node[lay] at (-6.7,-6.5) {L2};
  \node[lay] at (-6.7,-9.9) {L3};
\end{tikzpicture}}
\caption{The routing decision. L1 reduces the contract to an eight-dimensional
feature vector and a difficulty; L0 removes infeasible tracks before any scoring
and abstains rather than guessing when nothing is feasible; L2 scores the
surviving tracks under Equation~\ref{eq:mirt} and decides between exploiting the
best estimate and measuring an eligible contender; L3 records the outcome, which
moves the ability estimates that L2 will use next time.}
\label{fig:router-arch}
\end{figure}
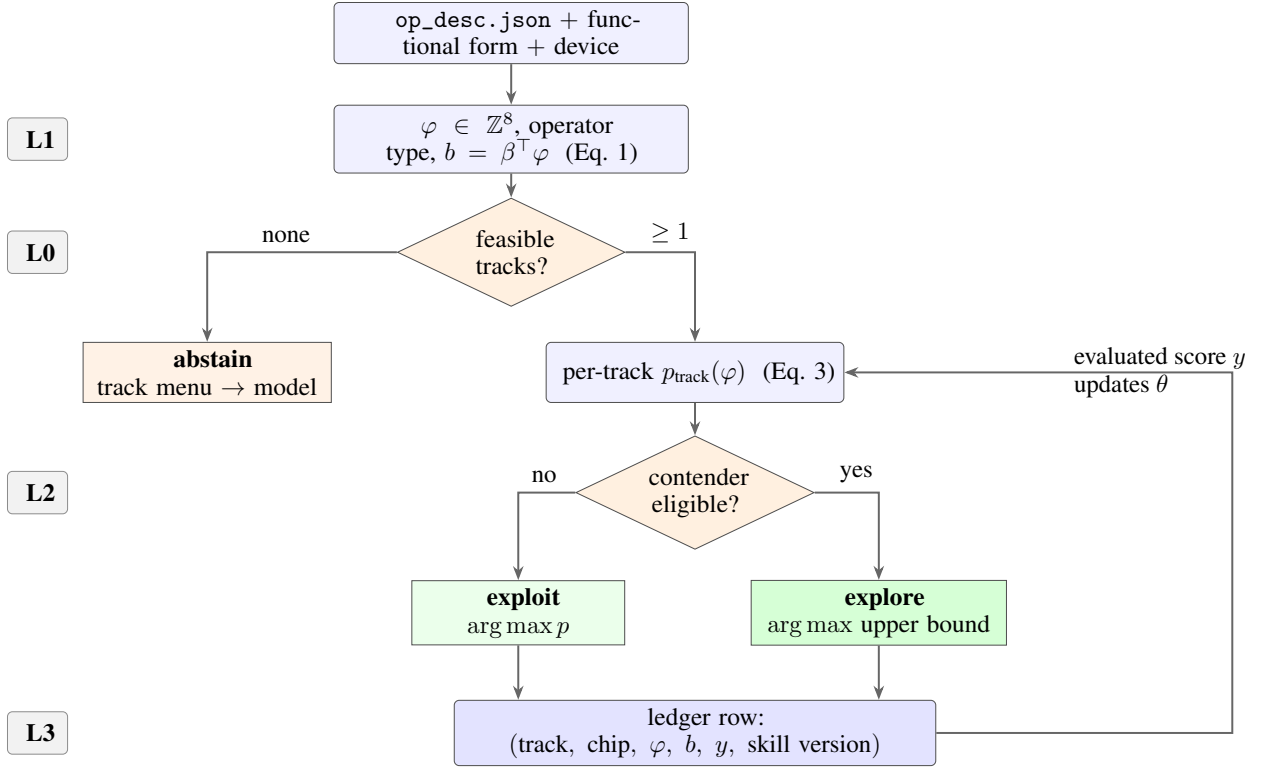

Track performance depends on both the operator's structure and the current
code-generation skills. A track that works well for one operator family may
perform poorly on another, and skill updates can change this relationship.
CAKE therefore estimates track suitability from previous evaluations. The
router has four components (Figure~\ref{fig:router-arch}):
a feasibility gate, a feature extractor, a scoring layer, and an append-only
observation ledger.

\paragraph{L0: feasibility gate.}
A declared capability table marks each \texttt{(track, chip, operator type)}
triple feasible or not, and infeasible tracks are removed before any scoring
happens. This prevents attempts on unsupported configurations. Forward binary
compatibility widens the declared set (a track feasible on 910B is feasible on
910C). On the 950, only DSL-C is
feasible for vector and cube--vector operators, so those operators land on DSL-C
without any scoring; and the SIMT track exclusively owns the SIMT operator type,
which no other track accepts, so the two never compete. When the gate leaves
nothing feasible, the router \emph{abstains}: it emits no
decision and hands the model the full track menu with each track's skills.
Abstention has two causes that are reported separately --- an unrecognized
device string, so that no gate can be applied at all, and a recognized device on
which the requested operator type has no feasible track.

\paragraph{L1: operator features and difficulty.}
The contract and the functional form are reduced to an eight-dimensional feature
vector $\varphi$ over the dimensions
$(T, S, B, R, D, M, F, C)$ --- tiling structure, scheduling depth, blocking,
reduction, dtype and quantization spread, data movement and irregular access,
fusion breadth, and cube intensity. Operators already characterized in the
library get an exact vector derived from their reference AST; new operators get a
heuristic vector, which is the normal case in production. A scalar difficulty
follows from a calibrated weight vector $\beta$,
\begin{equation}
  b \;=\; \beta^{\top}\varphi, \qquad
  \beta = (0.51,\,0.78,\,0.59,\,0.70,\,0.44,\,0.61,\,0.64,\,0.98),
  \label{eq:difficulty}
\end{equation}
pinned to a feature-specification version so that difficulty stays comparable
across recalibrations. Crucially $b$ is a property of the operator, not of the
track, which is what makes the routing decision robust to noise in a heuristic
$\varphi$: in the scalar regime below, $b$ cancels when tracks are compared.

\paragraph{L2: track ability and the routing score.}
Raw mean scores are difficult to compare when tracks have attempted different
operator sets. On
the current corpus DSL-A's mean achieved score is $0.576$ against DSL-C's
$0.471$, a gap that would read as a clear ordering; the ability estimates below
put them at $-0.95 \pm 0.38$ and $-0.99 \pm 0.36$, statistically
indistinguishable under this model. The raw $0.105$ difference alone therefore
does not establish a track-ability advantage.

CAKE therefore estimates a per-track ability $\theta$ under a continuous
item-response model. One observation is a completed generation-and-evaluation
run: a track attempted an operator of known difficulty $b$ and achieved a
normalized score $y \in [0,1]$. The link is
\begin{equation}
  \mathbb{E}[\,y \mid \theta, b\,] \;=\; \sigma(\theta - b), \qquad
  \sigma(x) = \bigl(1 + e^{-x}\bigr)^{-1},
  \label{eq:irt-link}
\end{equation}
with a beta likelihood on the continuous score and a posterior over $\theta$
obtained by expected-a-posteriori integration against an $\mathcal{N}(0,1)$
prior on a fixed grid. Because $b$ is computable from $\varphi$ before any
observation exists (Equation~\ref{eq:difficulty}), the router can assign an
initial score to a track with no history using the prior. Such a score reflects
prior assumptions rather than measured track performance.

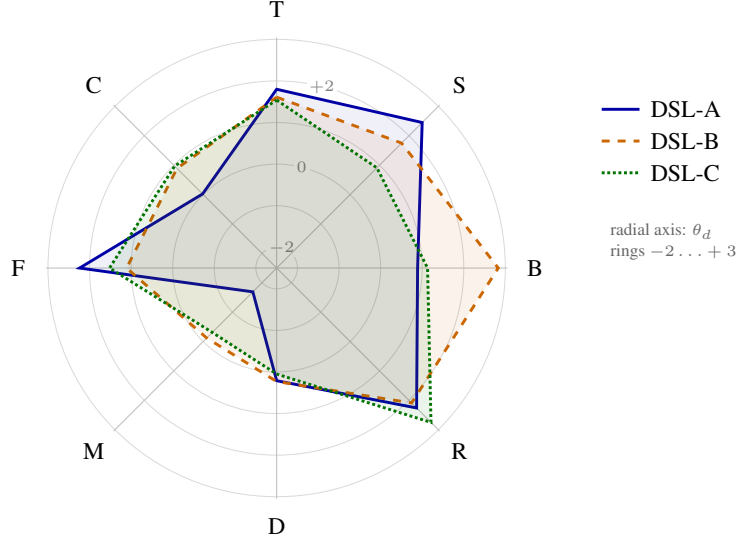
\begin{figure}[t]
\centering
\begin{tikzpicture}[scale=1.0,
    axline/.style={draw=black!25, line width=0.4pt},
    ring/.style={draw=black!15, line width=0.35pt},
    lbl/.style={font=\scriptsize},
    tick/.style={font=\fontsize{6pt}{7pt}\selectfont, text=black!50}]

  \foreach \t/\r in {-2/0.275, -1/0.825, 0/1.375, 1/1.925, 2/2.475, 3/3.025} {
    \draw[ring] (0,0) circle (\r);
  }
  \foreach \t/\r in {{$-2$}/0.275, {$0$}/1.375, {$+2$}/2.475} {
    \node[tick, fill=white, inner sep=1pt] at (76:\r) {\t};
  }
  \foreach \a/\nm in {90/T, 45/S, 0/B, 315/R, 270/D, 225/M, 180/F, 135/C} {
    \draw[axline] (0,0) -- (\a:3.05);
    \node[lbl] at (\a:3.42) {\nm};
  }

  \draw[draw=blue!65!black, line width=1.1pt, fill=blue!65!black, fill opacity=0.06]
    (90:2.365) -- (45:2.7225) -- (0:1.8645) -- (315:2.618) -- (270:1.4905)
    -- (225:0.4455) -- (180:2.607) -- (135:1.386) -- cycle;
  \draw[draw=orange!80!black, line width=1.1pt, dashed,
        fill=orange!80!black, fill opacity=0.08]
    (90:2.2605) -- (45:2.3375) -- (0:2.9315) -- (315:2.5245) -- (270:1.5015)
    -- (225:1.309) -- (180:1.9855) -- (135:1.8645) -- cycle;
  \draw[draw=green!45!black, line width=1.1pt, densely dotted,
        fill=green!45!black, fill opacity=0.08]
    (90:2.222) -- (45:1.8755) -- (0:1.9965) -- (315:2.8875) -- (270:1.4025)
    -- (225:1.2045) -- (180:2.211) -- (135:1.9085) -- cycle;

  \begin{scope}[shift={(4.3,2.1)}, font=\scriptsize]
    \draw[draw=blue!65!black, line width=1.1pt] (0,0) -- (0.5,0)
      node[right, black] {DSL-A};
    \draw[draw=orange!80!black, line width=1.1pt, dashed] (0,-0.42) -- (0.5,-0.42)
      node[right, black] {DSL-B};
    \draw[draw=green!45!black, line width=1.1pt, densely dotted] (0,-0.84) -- (0.5,-0.84)
      node[right, black] {DSL-C};
    \node[align=left, text=black!60, font=\fontsize{6.5pt}{8pt}\selectfont]
      at (0.95,-1.75)
      {radial axis: $\theta_d$\\rings $-2 \ldots +3$};
  \end{scope}
\end{tikzpicture}
\caption{Fitted per-dimension abilities $\theta_d$ for the three DSL tracks
under one backbone. Axes are the eight feature dimensions of
Section~\ref{subsec:dsl-routing}. The tracks are nearly indistinguishable on
tiling ($T$), reduction ($R$), and dtype spread ($D$), and separate on blocking
($B$, where DSL-B reaches $+2.83$), data movement ($M$, where DSL-A is $-1.69$
against DSL-C's $-0.31$), and scheduling depth ($S$, where DSL-A is $+2.45$
against DSL-C's $+0.91$). These gaps are what make the routing ordering
operator-dependent rather than fixed.}
\label{fig:track-radar}
\end{figure}

A single scalar $\theta$ per track is not enough. With scalar abilities the
comparison $\sigma(\theta_{t} - b)$ ranks tracks identically for every operator,
since $b$ is shared --- the router would degenerate into a fixed preference
order. CAKE therefore fits a per-dimension ability model per track: writing
$z = \operatorname{logit}(y) + b$, a ridge-regularized weighted least squares
(intercept unpenalized) gives
\begin{equation}
  z \;\approx\; a_0 \;+\; \sum_{d} \theta_d \cdot \frac{\varphi_d}{3},
  \qquad
  p_{\text{track}}(\varphi) \;=\; \sigma\!\Bigl(a_0 + \sum_{d} \theta_d \frac{\varphi_d}{3} - b\Bigr),
  \label{eq:mirt}
\end{equation}
where $a_0$ is the track's overall ability on the same scale as the scalar
$\theta$ and $\theta_d$ is its ability increment on dimension $d$. The
interaction terms $\theta_d \varphi_d$ are what make the ranking
operator-dependent. Fitting requires at least as many valid observations as free
parameters; below that threshold the router falls back to the scalar form and
records the fallback in its decision log. Figure~\ref{fig:track-radar} shows the fitted profiles: the
three DSL tracks are close on tiling, reduction, and dtype handling, and separate
sharply on data movement ($M$) and blocking ($B$).

\paragraph{A worked decision.}
Table~\ref{tab:router-decisions} traces four operators through
Equation~\ref{eq:mirt} on the 910B. The first pair illustrates how operator
features change the selected track.
\texttt{gelu} and \texttt{gather} are both vector operators with the same
feasible track set and similar difficulty ($b = -2.18$ against $-1.40$ after
centering). Their feature vectors differ in data movement, where $\varphi_M$
goes from $0$ to $2$, and dtype spread, where $\varphi_D$ goes from $2$ to $1$.
The larger between-track contrast is in data movement: DSL-A's $\theta_M$ is
$-1.69$ while DSL-C's is $-0.31$. As a result,
DSL-A's lead on \texttt{gelu} ($0.694$ against $0.613$) becomes a deficit on
\texttt{gather} ($0.239$ against $0.366$), and the decision flips. The remaining
two rows show two further dimensions taking control: \texttt{conv\_2d} has
$\varphi_B = 3$ and goes to DSL-B, whose $\theta_B = +2.83$ is the largest
single coefficient in the model; \texttt{mha} carries $\varphi_S = 3$
and returns to DSL-A on scheduling depth, where $\theta_S = +2.45$ against
DSL-C's $+0.91$ is the largest per-dimension difference in DSL-A's favour.
Reduction works the other way, since DSL-C's $\theta_R = +2.75$ exceeds
DSL-A's $+2.26$. DSL-A wins even though \texttt{mha}'s centered difficulty is
$+7.57$ and every track's predicted normalized score is low. The four examples select three
different tracks despite using the same device and fitted track profiles.

\paragraph{Evaluation ledger and skill versions.}
The observation ledger is append-only and grows one row per evaluated operator:
the track used, the device, the difficulty and feature vector, the achieved
normalized score, and a fingerprint of the code-generation skills that produced
it. Any evaluation harness that returns a per-operator score on a comparable
scale can supply $y$ --- CAKE-Bench's per-operator score is one such source, a
user's own metric is another. This allows new operator evaluations to update
the router without rerunning a fixed benchmark suite. Observations flagged as infrastructure
failures, as reward-hacking candidates, or as internally inconsistent
submissions are retained in the ledger but excluded from ability estimation.
To account for changes in code-generation skills, the router identifies each
track configuration by the pair
\texttt{(track, skill\_version)}: observations from a superseded skill version
enter estimation at half weight, which widens the posterior and prompts
re-measurement of the updated track. New evaluations receive full weight.

\paragraph{Selecting tracks for exploration.}
Testing a track requires a full generation, build, and evaluation cycle.
CAKE limits these additional evaluations using two criteria.
A Bernoulli-style information proxy $I = p(1-p)$ at the current estimate is the
\emph{eligibility} heuristic. For the continuous scores modeled here, this is
not the Fisher information of the beta likelihood. An
optimistic upper bound $\theta + 1.96\,\mathrm{se}$ identifies tracks that could
outperform the currently preferred track. Among eligible alternatives, the
router selects the one with the highest upper bound. Thus, the information
proxy controls eligibility, while the upper bound determines which alternative
to test. Ranking by the information proxy alone would instead favor scores
near $0.5$, where $p(1-p)$ is largest.

After a skill update, the router compares how many current-version evaluations
are available for each track. Tracks with fewer observations than their peers
receive a re-measurement bonus. If all tracks are equally under-sampled, no
differential bonus is applied; the ordinary eligibility and upper-bound criteria
still govern exploration.

\paragraph{The routing loop.}
Each completed evaluation updates the evidence used for subsequent routing. At the
routing stage the router writes its decision next to the operator's artifacts,
together with the full candidate snapshot --- every feasible track's ability
estimate, predicted normalized score, information proxy, and skill fingerprint. At the
evaluation stage the achieved score is written back as a new observation, the
affected track's posterior moves, and the next operator of similar character may
route differently. Both events are also appended to a central timeline, which
also records abstentions and routing errors. The saved features, scores, and
skill versions allow a past decision to be reconstructed using the information
available at the time.

\begin{table}[t]
  \centering
  \footnotesize
  \setlength{\tabcolsep}{5pt}
  \caption{Four routing decisions on the 910B under Equation~\ref{eq:mirt}
  (FrontierModel-A backbone, pure exploitation). $\varphi$ is printed in dimension order
  $(T,S,B,R,D,M,F,C)$ and $b$ is centered on the corpus mean. Columns
  $p_{\text{A}}$, $p_{\text{B}}$, $p_{\text{C}}$ are the predicted normalized scores of the
  three DSL tracks. The last column names the dimension whose ability
  coefficients decide the ordering.}
  \label{tab:router-decisions}
  \begin{tabular}{@{}l l r r r r c l@{}}
    \toprule
    \textbf{Operator} & \boldmath$\varphi$ & \boldmath$b$ &
    \boldmath$p_{\text{A}}$ & \boldmath$p_{\text{B}}$ & \boldmath$p_{\text{C}}$ &
    \textbf{Chosen} & \textbf{Decided by} \\
    \midrule
    \texttt{gelu}     & (0,0,1,0,2,0,0,0) & $-2.18$ & \textbf{0.694} & 0.463 & 0.613 & DSL-A & baseline ability \\
    \texttt{gather}   & (0,0,1,0,1,2,0,0) & $-1.40$ & 0.239 & 0.252 & \textbf{0.366} & DSL-C & $M$ (data movement) \\
    \texttt{conv\_2d} & (1,0,3,0,2,0,0,2) & $+1.47$ & 0.165 & \textbf{0.314} & 0.218 & DSL-B & $B$ (blocking) \\
    \texttt{mha}      & (3,3,3,3,2,0,1,2) & $+7.57$ & \textbf{0.257} & 0.168 & 0.101 & DSL-A & $S$ (scheduling depth) \\
    \bottomrule
  \end{tabular}
\end{table}

\newcommand{\ascendninefiftyappendix}{%
\section{Ascend 950 Vector: Regbase SIMD and SIMT}
\label{app:a5}
This appendix documents how CAKE generates and adapts operator code for the Ascend~950 series.
The 950 (\texttt{a5}, chip \texttt{C310}, \texttt{device\_type=950},
\texttt{\_\_DAV\_C310\_\_}) shares its \emph{Cube} primitive set with the
earlier a2/910B (\texttt{C220}); the divergence is on the \emph{vector} side.
Two 950-only paths live on that side: \emph{Regbase}, a register-level SIMD
path for latency-sensitive kernels with regular access patterns, and SIMT, a
multi-thread execution model for operators whose memory access or control flow
is data-dependent. Sections~\ref{app:a5-regmodel} and~\ref{app:a5-simtblock}
describe the hardware resources. Section~\ref{app:a5-path-selection} gives the
path-selection logic; Sections~\ref{app:a5-regbase} and~\ref{app:a5-simt}
describe generated kernels, validation signals, and the SIMT programming-model
catalogue.

\subsection{Register-level operand model}
\label{app:a5-regmodel}

Where the 910 path issues whole-tensor operations over UB-resident
\texttt{LocalTensor}s (\emph{membase}), the 950 exposes a register-level
\emph{MicroAPI} programmed through \texttt{@vf} micro-functions. The
hardware-specific code-generation choices are therefore:

\begin{itemize}
    \item \textbf{Register batches:} \texttt{RegTensor<T>} holds a fixed \texttt{VREG\_SIZE}-byte vector register rather than an arbitrarily large UB buffer.
    \item \textbf{Predication, not counts:} active lanes are carried by a \texttt{MaskReg} predicate passed to every op, replacing the membase \texttt{count} argument.
    \item \textbf{Explicit addressing:} gather/scatter index by raw byte offset with no bounds check. The regbase path offers \texttt{LoadUnAlign}/\texttt{StoreUnAlign} for UB$\leftrightarrow$register movement, enabling \emph{unaligned} access.
    \item \textbf{MicroAPI:} \texttt{add}/\texttt{sub}/\texttt{exp}/\texttt{log} exist on both paths but operate on registers in the 950 MicroAPI. Operations without a supported register-level lowering, such as widening \texttt{Mull} or reading back a \texttt{Squeeze} count, retain their membase implementation.
\end{itemize}

\begin{table}[tb]
\centering\small
\begin{tabular}{@{}lll@{}}
\toprule
& \textbf{Membase} & \textbf{Regbase} \\
\midrule
Residency  & \texttt{LocalTensor} in UB & \texttt{RegTensor} in VREG \\
API        & whole-tensor \texttt{AscendC::*} & \texttt{AscendC::Reg::*} / MicroAPI \\
Addressing & block-strided aligned & can be unaligned \\
Masking    & \texttt{count} argument & explicit \texttt{MaskReg} predicate \\
Movement   & \texttt{DataCopy} (TPipe queue) & \texttt{Load(Un)Align}/\texttt{Store(Un)Align} \\
\bottomrule
\end{tabular}
\caption{Membase and Regbase operand models: data residency, addressing,
masking, and transfer operations.}
\label{tab:mb-vs-rb}
\end{table}

\paragraph{How Regbase reshapes the compute ceiling.}
Regbase is an architectural shift from classical SIMD to a
\emph{dual-issue, register-based SIMD} design: \texttt{@vf} kernels run
micro-instructions directly on \texttt{RegTensor} operands held in the vector
register file. Whereas each membase vector operation reads from UB and writes
back, Regbase keeps the working set resident in registers across the
micro-kernel. Removing those UB round trips moves the throughput limit from UB
bandwidth toward instruction issue, widened by dual issue.

\subsection{SIMT block resources}
\label{app:a5-simtblock}

Each AI~Vector core exposes a SIMT block that runs up to 2\,048 threads
organized in 32-thread warps, with warp-level shuffle, vote, and reduction
intrinsics and 11-operation atomics on both on-chip and global memory. SIMT
threads access the \emph{DataCache} partition of the Unified Buffer---at most
128\,KB, with the remaining UB capacity reserved for the SIMD path---and
synchronize through block-wide barriers.

\subsection{Path selection}
\label{app:a5-path-selection}

The capability gate of Section~\ref{subsec:dsl-routing} first determines
whether a supported execution track exists. On the 950, data-dependent,
non-contiguous access selects SIMT; otherwise, register-level DSL coverage and
the known gap list select Regbase or the membase fallback.

\begin{figure}[tb]
\centering
\begin{tikzpicture}[node distance=0.9cm and 1.4cm]
    \node[proc] (in) {op\_desc + \texttt{check\_device.sh}};
    \node[dec, below=of in] (t) {target 910 or 950?};
    \node[term, below left=1.1cm and 0.3cm of t] (910) {\textbf{910}\\AscendC API};
    \node[proc, below right=1.1cm and 0.3cm of t] (950) {\textbf{950}};
    \node[dec, below=1.1cm of 950] (acc) {data-dependent\\non-contiguous access?};
    \node[proc, below left=1.1cm and 0.3cm of acc] (dsl) {DSL expresses\\kernel? (no gap)};
    \node[term, below right=1.2cm and -0.2cm of acc] (simt) {\textbf{SIMT}\\App.~\ref{app:a5-simt}: Q1--Q6};
    \node[term, below left=1.1cm and -0.2cm of dsl] (reg) {\textbf{Regbase}\\\texttt{@vf} MicroAPI codegen};
    \node[term, below right=1.1cm and -0.2cm of dsl] (mem) {\textbf{Membase}\\AscendC \texttt{LocalTensor}};
    \draw[flow] (in) -- (t);
    \draw[flow] (t) -| (910) node[pos=0.25, above, font=\scriptsize] {910B};
    \draw[flow] (t) -| (950) node[pos=0.25, above, font=\scriptsize] {950};
    \draw[flow] (950) -- (acc);
    \draw[flow] (acc) -| (dsl) node[pos=0.25, above, font=\scriptsize] {no};
    \draw[flow] (acc) -| (simt) node[pos=0.25, above, font=\scriptsize] {yes};
    \draw[flow] (dsl) -| (reg) node[pos=0.25, above, font=\scriptsize] {yes};
    \draw[flow] (dsl) -| (mem) node[pos=0.25, above, font=\scriptsize] {gap};
\end{tikzpicture}
\caption{Target and path routing. \texttt{check\_device.sh} reports the chip
generation, which admits the 910 lane or the two 950-specific lanes. On the
950, operators with data-dependent non-contiguous access take SIMT; the
remainder use Regbase when the register-level DSL covers the kernel and fall
back to membase when it does not. The SIMT model-selection questions Q1--Q6
are given in Appendix~\ref{app:a5-simt}.}
\label{fig:router}
\end{figure}

\subsection{Regbase code generation}
\label{app:a5-regbase}

The Regbase generation path uses explicit register operations to reduce UB
transfers or improve addressing and instruction scheduling.
Table~\ref{tab:mb-vs-rb} contrasts its operand model with membase.
CAKE uses the DSL-C lowering skills to generate this code.

    \paragraph{When the router selects Regbase.}
    Three signal classes trigger the path:
    (1)~\textbf{regbase-only micro-instructions} --- kernels using \texttt{DataCopyGather}, \texttt{Histograms}, and other 950 register-based micro-instructions that have no equivalent in the AscendC API; (2)~\textbf{unaligned UB access} --- kernels that require non-aligned access into the Unified Buffer; (3)~\textbf{low-latency} kernels --- small blocks that fit a single register and high-iteration inner loops. Membase remains the default; Regbase is attempted only when DSL-C supports all required operations.

    \paragraph{Generated code artifacts.}
    The Regbase path emits a \texttt{@vf} vector-function kernel with the
    structure in Figure~\ref{fig:vf}: register and mask allocation followed by
    \texttt{CeilDiv(len, VREG\_LANES)} loop iterations. The mask handles the
    final partial iteration. Reusable templates cover lookup-table/gather,
    compaction/filtering, and histogram/scatter operations. Post-processing
    places the generated code in CAKE's operator project and renames the entry
    point to match its binding. Migration from an existing membase kernel
    proceeds one region at a time, checking bit-exact agreement with the
    original implementation after each conversion. Regions without a supported
    register-level lowering remain in membase form.

    \begin{figure}[tb]
    \centering
    \begin{tikzpicture}[node distance=0.45cm]
        \node[phase, fill=blue!4] (decl) {\textbf{Declare}\;:\; \texttt{RegTensor<T>} (max 4/type), one \texttt{MaskReg}, \texttt{Var}: \texttt{one\_repeat=VREG\_LANES}, \texttt{repeat\_times}, \texttt{rem}};
        \node[phase, below=of decl] (mask) {\textbf{Update mask}\;:\; \texttt{update\_mask(m, rem)} --- active lanes shrink at the tail};
        \node[phase, below=of mask] (load) {\textbf{Load}\;:\; \texttt{LoadAlign} / \texttt{ub\_to\_reg} (UB$\to$reg) with on-load \texttt{LoadDist} (unpack / downsample / brcb)};
        \node[phase, below=of load] (comp) {\textbf{Compute (all masked by m)}\;:\; elementwise + transcendental; \texttt{Cast}/RoundMode; \texttt{compare}$\to$\texttt{MaskReg}$\to$\texttt{select}; cross-reg \texttt{reduce}; \texttt{PipeBarrier} only across pipes};
        \node[phase, below=of comp] (store) {\textbf{Store}\;:\; \texttt{StoreAlign} / \texttt{reg\_to\_ub} (reg$\to$UB, mask-gated) with on-store \texttt{StoreDist} (pack / interleave)};
        \node[below=0.35cm of store, font=\scriptsize, text=black!60] (dec) {\texttt{rem -= one\_repeat}; loop. No separate tail loop.};

        \draw[flow] (mask) -- (load);
        \draw[flow] (load) -- (comp);
        \draw[flow] (comp) -- (store);
        \draw[flow] (store) -- (dec);
        \begin{scope}[on background layer]
            \node[draw=black!40, dashed, rounded corners, fit=(mask)(store), inner sep=5pt, label={[font=\scriptsize, text=black!55]right:loop $\times$ \texttt{repeat\_times}}] (loopbox) {};
        \end{scope}
        \draw[flow] (decl) -- (mask);
    \end{tikzpicture}
    \caption{\texttt{@vf} Regbase kernel skeleton. In this pattern, narrow I/O
    (bf16/int8) is widened to \texttt{RegTensor<float>} on load via
    \texttt{LoadDist} and packed back on store; computation uses float registers.}
    \label{fig:vf}
    \end{figure}

    \paragraph{Compiler and profiler checks.}
    CAKE checks whether the conversion reduces memory traffic and execution
    time using the following signals:
    \begin{itemize}
        \item \textbf{PipeBarrier removal} --- same-pipe (V$\to$V) barriers disappear, freeing instruction/pipeline slots; barriers remain only across pipes (MTE2$\to$V, V$\to$MTE3).
        \item \textbf{Lower load/store ratio} --- register residency removes UB round-trips that membase incurs.
        \item \textbf{Fewer wall cycles} --- \texttt{cannsim} reports Regbase $<$ membase on the same case, with a per-pipe split identifying the dominant pipe.
    \end{itemize}
    These are read from \texttt{cannsim record ./test\_aclnnop -s Ascend950 --gen-report} and, for per-instruction attribution, \texttt{msprof op simulator} (per-instruction CSV mapped back to source lines).

\paragraph{Validation without 950 hardware.}
In the absence of 950 hardware, CAKE offers a simulator-only path backed by
\texttt{cannsim} or \texttt{msprof} that enables preliminary validation.

\subsection{SIMT code generation}
\label{app:a5-simt}
Membase and Regbase execute predicated vector instructions over UB-resident or register-resident data. Some operators, including gather, histogram, and top-k selection, involve data-dependent addresses or control flow. Such patterns are not inherently inexpressible in SIMD: supported gather/scatter operations, masking, and staged data movement can implement many of them. However, some irregular patterns require additional transfers, predication, or serial processing, depending on the available APIs and lowering support. SIMT provides an alternative implementation path for these cases.

The 950 SIMT execution model can reduce these overheads using the block resources
summarized in Section~\ref{app:a5-simtblock}. The capability gate of
Section~\ref{subsec:dsl-routing} determines whether an operator can enter the
SIMT track at all. The path is taken when two additional conditions hold beyond
the Regbase/membase split:
(i)~the operator exhibits irregular or data-dependent memory access, and
(ii)~the irregular access is \emph{non-contiguous}---contiguous scans with
data-dependent predicates can remain on the SIMD path when supported by the available operations.  Once on the
SIMT path, a six-question decision tree (Figure~\ref{fig:simt-decision})
refines the choice among three programming models that trade generality for
co-processor use. The simt-codegen skill implements this selection and generates
the corresponding kernel structure.

\begin{figure}[htbp]
\centering
\begin{tikzpicture}[
    node distance=0.7cm and 1.0cm,
    q/.style={diamond, aspect=2.2, draw=black!65, fill=orange!10,
              align=center, font=\scriptsize, inner sep=1pt,
              text width=2.6cm},
    m/.style={rectangle, rounded corners=2pt, draw=black!65,
              align=center, font=\scriptsize, minimum height=0.7cm,
              inner sep=3pt},
    mA/.style={m, fill=memC!12, text width=2.4cm},
    mB/.style={m, fill=anaC!12, text width=2.4cm},
    mC/.style={m, fill=evoC!12, text width=2.4cm},
    mV/.style={m, fill=black!6, text width=2.0cm},
    arr/.style={-{Stealth[length=1.6mm]}, draw=black!55, thick},
    lbl/.style={font=\scriptsize\itshape, text=black!60},
]
    \node[q] (q1) {Q1: irregular /\\data-dep access?};
    \node[mV, left=1.6cm of q1] (vec1) {Vector path};
    \draw[arr] (q1) -- (vec1) node[midway, above, lbl] {no};

    \node[q, below=0.8cm of q1] (q2) {Q2: input access\\contiguous?};
    \draw[arr] (q1) -- (q2) node[midway, right, lbl] {yes};
    \node[mV, left=1.6cm of q2] (vec2) {Vector path\\(SIMD+mask)};
    \draw[arr] (q2) -- (vec2) node[midway, above, lbl] {yes};

    \node[q, below=0.8cm of q2] (q3) {Q3: vectorizable\\compute fused?};
    \draw[arr] (q2) -- (q3) node[midway, right, lbl] {no};

    \node[q, right=1.8cm of q3] (q4) {Q4: purely\\elementwise?};
    \draw[arr] (q3) -- (q4) node[midway, above, lbl] {yes};
    \node[mC, below right=0.6cm and -0.4cm of q4] (mc) {\textbf{Model C}\\SIMT$\to$SIMD};
    \node[mC, below left=0.6cm and -0.4cm of q4] (mcv) {\textbf{Model C}\\reductions in entry};
    \draw[arr] (q4) -- (mc) node[midway, right, lbl] {yes};
    \draw[arr] (q4) -- (mcv) node[midway, left, lbl] {no};

    \node[q, below=0.8cm of q3] (q5) {Q5: cross-thread\\cooperation?};
    \draw[arr] (q3) -- (q5) node[midway, right, lbl] {no};
    \node[mB, left=1.6cm of q5] (mb) {\textbf{Model B}\\per-thread};
    \draw[arr] (q5) -- (mb) node[midway, above, lbl] {no};

    \node[q, below=0.8cm of q5] (q6) {Q6: cooperation\\kind?};
    \draw[arr] (q5) -- (q6) node[midway, right, lbl] {yes};

    \node[mB, below left=0.5cm and 0.0cm of q6] (mb1) {\textbf{B}+warp shuffle\\(reduction)};
    \node[mB, below=0.5cm of q6] (mb2) {\textbf{B}+atomics\\(scatter-accum)};
    \node[mA, below right=0.5cm and 0.0cm of q6] (ma) {\textbf{A} pure SIMT\\(compaction)};
    \draw[arr] (q6) -- (mb1);
    \draw[arr] (q6) -- (mb2);
    \draw[arr] (q6) -- (ma);
\end{tikzpicture}
\caption{SIMT model selection.  Questions Q1--Q2 gate entry to the SIMT
  path; Q3--Q6 choose among three programming models.  Model~A is a
  CUDA-like \texttt{\_\_global\_\_} kernel; Model~B bridges SIMT threads
  into the AscendC entry via \texttt{asc\_vf\_call}; Model~C adds a
  \texttt{\_\_simd\_vf\_\_} stage that offloads vectorizable arithmetic
  to the SIMD pipe after SIMT has performed the irregular access.}
\label{fig:simt-decision}
\end{figure}

\paragraph{Three programming models.}
\textbf{Model~A} (\emph{pure SIMT}) is the most general: a single
\texttt{\_\_global\_\_} kernel launched with four-argument
\texttt{<<<blocks, threads, ubuf, stream>>>} syntax, compiled with the
\texttt{-cce-enable-puresimt} flag.  It supports templates, struct arguments,
and the full SIMT intrinsic set (warp shuffle, vote, atomics, cache hints)
but cannot call the AscendC Micro~API (\texttt{LoadAlign}, \texttt{DataCopy},
etc.).  This model suits operators whose compute is entirely irregular---histograms,
hash-table insertion, stream compaction---and whose output can be written
directly to global memory.

\textbf{Model~B} (\emph{SIMT~VF bridge}) wraps the per-thread compute in a
\texttt{\_\_simt\_vf\_\_} free function dispatched from an
\texttt{\_\_aicore\_\_} entry via \texttt{asc\_vf\_call}.  The three-argument
launch integrates with \texttt{msopgen} project scaffolding (tiling macros,
\texttt{REGISTER\_TILING\_DEFAULT}), making it the default for operators that
fit the standard CANN host/device contract---gather, cumulative sum, top-$k$
with warp reductions.

\textbf{Model~C} (\emph{hybrid SIMT+SIMD}) extends Model~B for operators
that fuse an irregular access phase with a regular vectorizable phase.  A
\texttt{\_\_simt\_vf\_\_} function performs the data-dependent read and
stages results into the DataCache; a \texttt{\_\_simd\_vf\_\_} function
then runs the same SIMD vector pipe used by the Regbase path (adds, relu,
exp) over the staged data; finally, \texttt{DataCopy} moves the result
to global memory.  The benefit of offloading arithmetic to SIMD depends on the workload and must be evaluated together with data-staging and synchronization costs; low SIMD utilization alone does not establish negligible overhead.  A variant of Model~C
handles fused reductions (softmax, layer normalization): since
\texttt{\_\_simd\_vf\_\_} cannot compile cross-element reductions
(\texttt{ReduceMax}, \texttt{ReduceSum}), these are placed in the
\texttt{\_\_aicore\_\_} entry function using high-level AscendC APIs.

}%

\section{Evolutionary Optimization}
\label{sec:evolution}
Generation mode provides a working implementation that can serve as the starting
point for performance optimization. Evolution mode, or CAKE-evo, explores tiling,
buffering, and fusion choices through parallel candidate generation and on-device
evaluation. Profiling identifies the bottlenecks to target in each round.

    \subsection{Profiling-driven analysis layer}

    Before generating variants, CAKE-evo estimates a performance ceiling from
    the input shapes and dtypes, device clock, HBM/L2 bandwidth, and peak
    cube/vector throughput. Following a roofline decomposition, the modeled
    lower-bound execution time is the maximum of four pipeline times,
    \begin{equation}
        T_{\min} \;=\; \max\bigl(t_{\text{hbm\_read}},\; t_{\text{write}},\; t_{\text{cube}},\; t_{\text{vector}}\bigr),
        \label{eq:roofline}
    \end{equation}

    where reads and writes are treated as separate channels that overlap. For
    $b_{\text{out}}$ output bytes, writes follow an L2-fill-first model: traffic
    up to the L2 capacity $C_{L2}$ uses bandwidth $\beta_{L2}$, and the remainder
    uses HBM bandwidth $\beta_{\text{HBM}}$:
    \begin{equation}
        t_{\text{write}}=
        \begin{cases}
        b_{\text{out}}/\beta_{L2} & b_{\text{out}}\le C_{L2}\\[2pt]
        C_{L2}/\beta_{L2} + (b_{\text{out}}-C_{L2})/\beta_{\text{HBM}} & b_{\text{out}}> C_{L2}.
        \end{cases}
    \end{equation}
    The dominant term in Eq.~\eqref{eq:roofline} is the \emph{bottleneck signature} (memory vs.\ compute, and which pipe), and the arithmetic intensity $\mathrm{AI}=\text{FLOPs}/\text{bytes}$ versus the ridge point $\beta_{\text{HBM}}^{-1}\!\cdot\!\text{peak}$ confirms the regime. A complementary \emph{fusion analysis} parses the operator's dataflow graph and flags intermediates that would otherwise be materialized to HBM, quantifying the traffic a loop fusion would save. Together these yield, per case: (i) a \emph{ceiling-derived target} $T_{\min}$ that replaces a hand-picked speedup goal, (ii) a bottleneck label that selects which optimization family to try, and (iii) a loop fusion recommendation.

    \begin{figure}[t]
    \centering
    \begin{tikzpicture}[x=1.0cm,y=0.8cm]
    \draw[fl] (0,0) -- (8.4,0) node[right,font=\scriptsize]{$\;\log(\text{FLOP/byte})$};
    \draw[fl] (0,0) -- (0,5.0) node[above,font=\scriptsize]{attainable perf $\log$};
    \draw[memC,very thick] (0.2,0.2) -- (4,4);
    \draw[cmpC,very thick] (4,4) -- (8,4);
    \node[cmpC,font=\scriptsize,anchor=south] at (6.0,4.12) {compute peak};
    \node[memC,font=\scriptsize,rotate=39] at (1.55,2.25) {memory roof (BW)};
    \filldraw (4,4) circle (1.6pt);
    \node[font=\scriptsize,anchor=south] at (4,4.12) {ridge};
    \draw[dashed,black!40] (4,0)--(4,4);
    \node[font=\scriptsize,anchor=north] at (4,-0.05){$\mathrm{AI}^\ast$};
    \filldraw[memC] (1.7,1.0) circle (1.8pt);
    \node[memC,font=\scriptsize,anchor=north] at (1.9,1.00) {vector op};
    \draw[fl,memC] (1.7,1.05) -- (1.7,1.62) node[midway,right=1pt,font=\scriptsize,black]{headroom};
    \filldraw[cmpC] (6.0,3.0) circle (1.8pt);
    \node[cmpC,font=\scriptsize,anchor=north] at (6.0,3.0) {matmul};
    \draw[fl,cmpC] (6.0,3.05) -- (6.0,3.92) node[midway,right=1pt,font=\scriptsize,black]{headroom};
    \node[memC!80!black,font=\scriptsize] at (1.5,0.35){memory-bound};
    \node[cmpC!80!black,font=\scriptsize] at (6.4,0.35){compute-bound};
    \end{tikzpicture}
    \caption{Roofline ceiling model (Eq.~\ref{eq:roofline}). Each case is placed by its arithmetic intensity (AI); the binding pipeline gives the roof $T_{\min}$, and the \emph{headroom} (gap from the measured point up to the roof) bounds how much evolution can still gain. A variant that reaches its roof ($\text{headroom}\!\approx\!1$) is converged.}
    \label{fig:roofline}
    \end{figure}

    The resulting $T_{\min}$ is an optimistic analytical floor under the assumed
    overlap and traffic model, not a measured attainable latency. Dependencies,
    contention, and instruction overhead can keep a correct kernel above this
    floor even when further optimization is ineffective. We define
    \emph{headroom} as the measured runtime divided by $T_{\min}$; a value near
    one indicates little remaining improvement under this model.

    \subsection{The evolutionary loop}

Each round (Fig.~\ref{fig:loop}) generates $N$ variants in parallel, assigning
each an optimization direction based on the identified bottleneck. For example,
a memory-bound chain may benefit from fusion, while a compute-bound kernel may
benefit from different tiling or instruction scheduling. Candidates are compiled,
evaluated, ranked by speedup, and grouped into good, medium, and poor tiers.
The next round samples implementations from these tiers as starting points.

During advisor-in-the-loop consultation (AIC), the Advisor examines the profile
of the best candidate and supplies bottleneck-specific suggestions for the next
round. The search stops when the measured runtime approaches the modeled floor
($\text{headroom}\!\approx\!1$, Fig.~\ref{fig:roofline}) or the round budget
is exhausted. Post-evolution mining (PMR) then extracts validated optimizations
as candidate rules for reuse on other operators.

    \begin{figure}[t]
    \centering
    \begin{tikzpicture}[node distance=0.5cm and 1.0cm]
    \node[ana] (an) {\textbf{Analyze}\\roofline $T_{\min}$,\\bottleneck, fusion};
    \node[evo, right=1.15cm of an] (sp) {\textbf{Spawn $N$ variants}\\distinct strategy\\per bottleneck};
    \node[evo, right=1.15cm of sp] (ev) {\textbf{Evaluate}\\compile + profile};
    \node[evo, below=0.95cm of ev] (ti) {\textbf{Tier}\\good / medium / poor};
    \node[ana, left=1.15cm of ti] (ai) {\textbf{AIC}\\advisor consults\\top-1 profiling};
    \node[dec, left=1.15cm of ai] (dc) {headroom\\$\approx 1$?};
    \node[ana, below=0.9cm of dc] (pm) {\textbf{Finalize + PMR}\\mine effective\\opts $\to$ rules};

    \draw[fl] (an) -- (sp);
    \draw[fl] (sp) -- (ev);
    \draw[fl] (ev) -- (ti);
    \draw[fl] (ti) -- (ai);
    \draw[fl] (ai) -- (dc);
    \draw[fl] (dc) -- node[left,font=\scriptsize]{yes} (pm);

    \coordinate (topY) at ($(an.north)+(0,0.6)$);
    \coordinate (nL)   at ($(dc.west)+(-0.55,0)$);
    \coordinate (nTL)  at (nL |- topY);
    \coordinate (nTR)  at (sp.north |- topY);
    \draw[fl] (dc.west) -- (nL) -- (nTL) -- (nTR) -- (sp.north);
    \node[above,font=\scriptsize] at ($(nTL)!0.5!(nTR)$)
        {no: inject elite + tier inspirations};

    \draw[fl,anaC,dashed] (an.south) -| (dc.north)
        node[pos=0.75,right,font=\scriptsize,black]{$T_{\min}$};
    \end{tikzpicture}
    \caption{The CAKE-evo loop. The analysis layer (blue) sets the ceiling target and bottleneck-driven variant assignment; the evolutionary stages (green) spawn, measure, and tier variants; AIC injects profiling-grounded guidance; the roofline headroom provides the stop test; PMR compounds knowledge into rules.}
    \label{fig:loop}
    \end{figure}

    \subsection{Strategy knowledge}

Optimization directions are drawn from a curated strategy library derived from
production operators and organized into five classes (Table~\ref{tab:classes}).
The kernel-level classes cover dtype handling (D), performance (P), and numerical
accuracy (A). Fusion strategies (F) eliminate intermediate global-memory
transfers; host-level strategies (H) select specialized kernels and distribute
work across cores according to input shapes.

    \begin{table}[tb]
    \centering\small
    \begin{tabular}{@{}llp{7.4cm}@{}}
    \toprule
    Class & Scope & Representative strategies \\
    \midrule
    D \,Multi-dtype   & kernel    & mixed precision; template / TilingKey dtype dispatch \\
    P \,Performance   & kernel    & double buffering; adaptive tiling; pipeline sync; 32B alignment; vectorized copy; matmul scheduling; L1/L2 cache hierarchy \\
    A \,Accuracy      & kernel    & FP32 intermediate; Welford mean/var; rounding mode; softmax stability; high-precision rsqrt \\
    \textbf{F} \,Fusion    & cross-op & elementwise-chain anti-spill; norm$+$epilogue; matmul$+$epilogue (CV) \\
    \textbf{H} \,Host-level & host    & TilingKey priority-chain dispatch; multi-core load balance; L2/L1 memory tiering \\
    \bottomrule
    \end{tabular}
    \caption{Strategy classes. D/P/A are kernel-level; F and H raise the altitude to cross-operator fusion and host-side, shape-driven dispatch.}
    \label{tab:classes}
    \end{table}

Table~\ref{tab:routing} maps profiling results to candidate strategies. Inputs
include the dominant pipeline in Eq.~\ref{eq:roofline}, opportunities to fuse
intermediate operations, and differences in bottlenecks across input shapes.
Assigning variants different applicable strategies lets a round test several
responses to the measured bottleneck.

    \begin{table}[tb]
    \centering\small
    \begin{tabular}{@{}lp{8.3cm}@{}}
    \toprule
    Bottleneck signature & Routed strategies \\
    \midrule
    memory-bound $+$ fusable chain   & Fusion to keep intermediates off GMEM; double buffering; alignment; vectorized copy; memory-tier selection \\
    compute-bound (vector)           & Adaptive tiling; small-dimension row merging; algorithm selection; FP32 intermediates where precision requires them \\
    compute-bound (cube)             & Matmul scheduling (SplitK/StreamK); L1/L2 cache optimization; matmul-epilogue fusion \\
    instruction / scalar-bound       & Algorithm redesign; specialized AscendC APIs; narrow synchronization \\
    load imbalance (straggler core)  & Host-side dispatch and multi-core load balancing \\
    shape regimes diverge across cases & Multiple TilingKeys with a specialized kernel per shape regime \\
    \bottomrule
    \end{tabular}
    \caption{Bottleneck-signature $\rightarrow$ strategy routing across the classes in Table~\ref{tab:classes}. The roofline verdict and fusion analysis select which optimization family each variant explores.}
    \label{tab:routing}
    \end{table}

    Finally, the library is not static: the PMR step (Fig.~\ref{fig:loop}) distills the optimizations that were \emph{validated} to improve performance during evolution into reusable rules, so each optimized kernel improves both the strategy set and the routing table for the next.

\subsection{Case study: evolving EngramGateFusion without an official kernel baseline}
\label{subsec:engram-evolution}

EngramGateFusion is a CANN-Bench level-3 operator that combines a per-step
gated recurrence with a dilated depthwise convolution. The benchmark specifies
the mathematical contract and a correctness reference, but provides no official
optimized kernel. A conventional speedup against a vendor implementation is
therefore unavailable. We instead use three anchors: CAKE-evo's first working
kernel, the unfused PyTorch composition (roughly one hundred small-operator
launches), and the case-specific hardware lower bound $T_{\min}$ from
Eq.~\eqref{eq:roofline}. The campaign covers 20 shape configurations and two
bf16 outputs; the final package passes all 20 official grading cases. Development
and round-to-round timing used Ascend910B2, while official grading used
Ascend 910C.

\begin{figure}[tb]
  \centering
  \includegraphics[width=\linewidth]{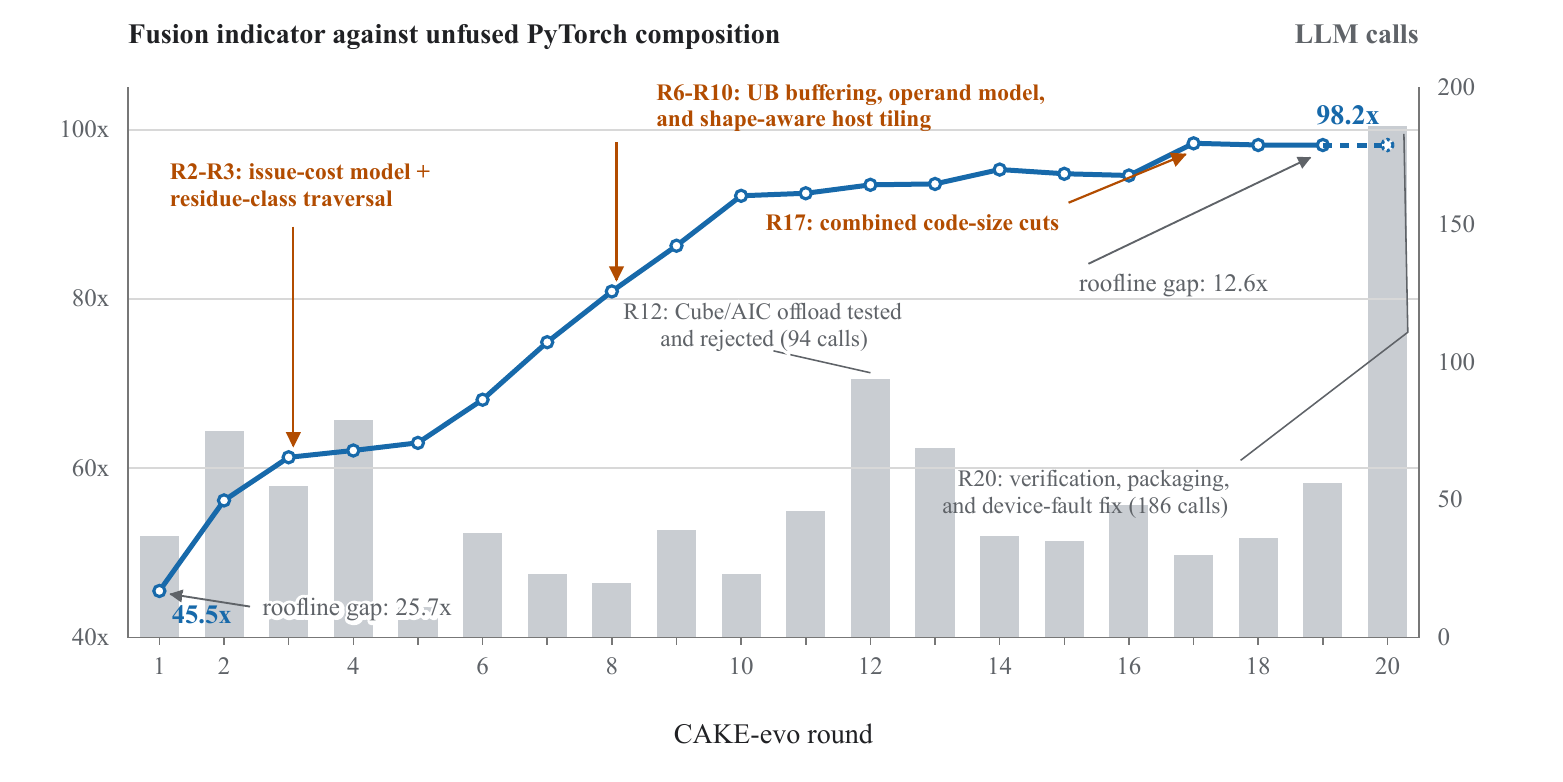}
  \caption{EngramGateFusion evolution over 20 CAKE-evo rounds and 20
  CANN-Bench cases. The line is the mean fusion indicator against the unfused
  PyTorch composition (left axis); gray bars show LLM calls in each round
  (right axis), and warm-colored callouts identify the mechanisms behind the
  largest gains. The indicator rises from $45.5\times$ to $98.2\times$, while
  the endpoint annotations show the mean kernel-to-roofline ratio falling from
  $25.7\times$ to $12.6\times$. Because the line mixes eager wall time with
  device kernel time, it is used only as a trajectory indicator; a separate
  same-basis measurement gives a final mean end-to-end speedup of $29.5\times$
  (geometric mean $27.6\times$). Round 20 performed verification, packaging,
  and a device-fault fix and therefore has calls but no new performance point;
  the dashed segment repeats the round-19 value and does not represent a new
  timing measurement.}
  \label{fig:engram-evolution}
\end{figure}

For round $r$ and case $i$, the campaign reports the fusion indicator
$S_r=\frac{1}{20}\sum_i t_{\mathrm{eager},i}/t_{r,i}$ and the roofline gap
$G_r=\frac{1}{20}\sum_i t_{r,i}/T_{\min,i}$. A larger $S_r$ indicates that the
fused kernel increasingly amortizes the unfused composition, while a smaller
$G_r$ indicates movement toward the hardware floor. The two ratios have
different evidentiary roles: $G_r$ compares device times directly, whereas
$S_r$ divides end-to-end eager wall time by device-only kernel time and is thus
a round-to-round fusion indicator, not a same-basis end-to-end speedup.
Figure~\ref{fig:engram-evolution} shows the evolution trajectory, with endpoint
roofline gaps, per-round LLM calls, and annotations of the main optimization
steps.

Across the 20 cases, aggregate device time falls from $1451.7\,\mu$s for the
round-1 kernel to $775.7\,\mu$s in round 19, a $46.6\%$ reduction or
$1.87\times$ within-system speedup. The local score rises from 64.53 to 72.80;
the packaged kernel scores 71.30 on the official grader with all 20 cases
passing. The precision report records bit-identical agreement on six cases and
at most one bf16 unit in the last place (ULP) on the remainder. The timing
improvement is measured against CAKE's round-1 kernel on the same development
device. The campaign reached its 20-round budget, not the roofline
stopping threshold.

The local and official scores are not directly comparable: the development
device has 48 vector cores and the grading device has 40, which also changes
shape-dependent dispatch. Round acceptance used a local oracle rather than the
official grading path. The first official submission failed 12 cases because
host-side UB accounting diverged from the kernel layout on the grading device.
Round 20 fixed this by deriving both layouts from a shared header; the final
20/20 result refers to the repaired package, not every intermediate candidate.

The performance gains followed several revisions to the agent's bottleneck
analysis. In round 2, measurements showed that the fixed issue overhead of a
vector API call dominated its length-dependent cost, making call reduction a
priority. In round 3, processing sequence positions by their index modulo the
dilation factor reduced the convolution ring buffer from $K$ times the dilation
factor to $K$ rows, where $K$ is the number of convolution taps.

Middle rounds introduced double buffering in the Unified Buffer (UB) and
shape-aware host tiling. The agent also revised its interpretation of
three-source vector instructions: replacing a chain of simpler operations with
one multiply-add could reduce instruction issue overhead. Later rounds tested
combinations of changes and revised a code-size rule that had rejected useful
candidates. In round 19, folding \texttt{conv\_norm\_weight} into the convolution
tap weights removed a multiplication from each row. The complete campaign used 36 generated variants
and 1,037 LLM calls across 20 rounds; 186 of those calls belong to the final
verification-and-packaging round. The resulting measurements and successful
changes provide material for post-evolution mining (PMR): instruction-cost
estimates, conditions under which a rule fails, and combinations of optimizations
that can be tested on other operators.

\section{CAKE Book and System-Level Self-Evolution}
\label{sec:self-evolving-agent}

Evolution mode improves one operator implementation by searching over
candidate variants under a fixed operator contract. CAKE also contains a
system-level improvement loop whose target is the generator itself. This loop is
implemented by two cooperating components: CAKE Book, an LLM Wiki-based
agentic knowledge-management system, and the Self-Evolving Agent, a persistent
AI agent that monitors CAKE runs and turns observed engineering experience into
human-reviewed improvements. Together they make CAKE less dependent on a
single model's prior knowledge: compilation failures, profiling signals,
successful optimizations, and human corrections become reusable system memory.

\subsection{CAKE Book as Agent-Native Reference}
\label{subsec:cake-book}

CAKE Book is the knowledge layer used by CAKE agents before, during, and after
operator generation. Following the open-source LLM Wiki pattern, CAKE Book
maintains wiki-style pages with bidirectional links, source metadata,
hardware and CANN-version tags, trust and recency weights, and an error
notebook for recurring Ascend/CANN development failures. Human review remains
available for proposed knowledge and repository updates. The design enables
knowledge reuse in later CAKE runs; its causal effect on success and performance
is not isolated by the current evaluation (Section~\ref{subsec:cake-book-evidence}).

CAKE Book complements the static references packaged with individual skills.
Static skill references capture stable procedures such as artifact schemas,
prompt contracts, code templates, and evaluation commands. CAKE Book captures
knowledge that changes with experience: version-specific compiler behavior,
newly observed CANN packaging traps, operator-family tiling lessons, 950
Regbase/SIMT notes, profiler interpretations, and workarounds that have only
recently been validated. When a skill needs such context, it can query CAKE Book
as a task-time reference source. The returned context is ranked not only by
semantic similarity, but also by link structure, evidence quality, hardware and
software compatibility, freshness, and human review status.

CAKE Book supports both retrieval and updates. For retrieval, an agent or
skill submits the current task context, retrieves relevant wiki pages and
linked error notes, ranks them by evidence, and packs the result into an
agent-ready reference. For updates, a completed run contributes new
observations: a compiler error and its fix, a profiler bottleneck and the code
change that removed it, a DSL route that repeatedly failed for an operator
family, or a human review note that corrects an agent assumption. CAKE Book
deduplicates these observations, attaches provenance, and exposes them for
review before they become trusted references for future agents.

\subsection{Run Evidence from the CAKE CI Platform}
\label{subsec:cake-ci-platform}

The Self-Evolving Agent observes CAKE through the CAKE CI Platform. The platform
standardizes generation and evolution runs and exposes a CLI through which an
agent can inspect run status, logs, and artifacts. For each run, the agent can
retrieve the operator contract, generated PyTorch reference, functional form,
DSL baseline, AscendC project files, build logs, CANN package and install logs,
\texttt{evaluate.py} outputs, profiling CSVs, skill traces, \texttt{PROGRESS.md},
evolution reports, Advisor state, issue or pull-request links, and user
annotations. These artifacts preserve both successful paths and failed attempts,
which is essential because many reusable CAKE lessons are negative: which route
compiled but failed precision, which template broke on non-aligned shapes, or
which Advisor suggestion did not transfer to a neighboring operator family.

The CI interface also makes self-evolution measurable. Runs can be grouped by
operator family, DSL route, target chip, CANN version, input mode, evaluation
case type, and final status. This lets CAKE Book distinguish local anecdotes
from repeated patterns and lets the Self-Evolving Agent detect regressions after
router, template, Advisor, or reference changes. Linking an update to its source
runs also allows later evaluations to check whether the same failure recurs.

\subsection{Self-Evolving Agent}
\label{subsec:self-evolving-agent}

The Self-Evolving Agent runs continuously, checking newly available CAKE CI
Platform runs. It reads
the corresponding artifacts, and records observations through CAKE Book. Its
daily work resembles an engineer's maintenance loop: summarize recent CI
successes and failures, identify repeated failure modes, compare them with
existing CAKE Book pages, notice when a previously useful rule becomes stale,
and identify cases where a dynamic lesson should become part of the repository.

When the observation is purely knowledge-oriented, the Self-Evolving Agent
interacts with CAKE Book directly. For example, it can add a new error-note page
for a CANN compiler diagnostic, link it to the affected operator family and CANN
version, attach the run IDs that demonstrate the workaround, and mark the page
as pending human review. When the observation implies a code or repository
change, the agent opens an issue or pull request against the CAKE repository.
Typical pull requests include adding a static skill reference, improving a DSL
prompt, tightening a lowering check, documenting a new Advisor rule, extending
an evaluation audit, or updating router guidance for a hardware-specific path.
Merging remains a human decision; the agent proposes changes and attaches
evidence, while maintainers decide whether the evidence is sufficient.

This separation keeps the self-evolving loop auditable. CAKE Book can absorb
fresh, uncertain, or version-specific experience quickly without silently
rewriting the public pipeline. Repository changes are slower and require review,
but once merged they improve the baseline behavior of all later agents. In this
sense, CAKE Book is the fast memory of the system, while the static skill
references, router rules, templates, and Advisor rules are the stabilized memory.

\subsection{Reference Promotion and Governance}
\label{subsec:reference-promotion}

Reference promotion is the bridge between dynamic and static knowledge. A CAKE
Book entry becomes a candidate for promotion when it is repeatedly useful,
supported by concrete run evidence, scoped to clear hardware and software
versions, and stable enough that future agents should see it without issuing a
live CAKE Book query. The Self-Evolving Agent prepares the promotion by linking
successful and failing runs, summarizing the lesson, identifying the affected
skill or rule file, and opening an issue or pull request. Human reviewers then
check whether the proposed reference is correct, sufficiently general, and safe
to apply across future runs.

Governance is especially important because Ascend/CANN behavior changes across
hardware and software versions. CAKE Book entries therefore carry provenance:
run IDs, operator families, target chip, CANN and torch-npu versions, input
shapes, pass/fail status, profiler files, and reviewer notes. Conflicting
evidence does not overwrite older knowledge; it creates versioned or scoped
pages. Stale-rule detection is handled in the same evidence-backed way: when CI
runs show that a promoted rule no longer helps or begins to harm a route, the
Self-Evolving Agent records the contradiction in CAKE Book and opens a follow-up
issue or pull request for human review.

Together, these components support knowledge reuse across runs. CAKE agents retrieve
current lessons from CAKE Book during generation, CAKE CI Platform captures the
full evidence trail of their attempts, the Self-Evolving Agent monitors that
trail around the clock, and stable lessons move back into static skill
references through reviewed pull requests. Operator-level optimization supplies
the observations; repository review determines which lessons become part of the
shared generation workflow.

\section{Evidence from CANN-Bench and Operator Campaigns}
\label{sec:results}

\paragraph{Setup.}
The evaluation target is the official CANN-Bench v1.1 task set on Ascend
950PR, comprising 53 operators across difficulty levels 1--4, with 20 standard
and 80 hidden cases per operator. We use the October 15, 2026 leaderboard
snapshot, which selects the best result for each operator separately in each
case set across benchmark revisions 1.1.0--1.1.2. The per-operator
\emph{benchsite score} is out of 100, with compile, functional-correctness,
and performance weights of $0.2$, $0.3$, and $0.5$ under
\texttt{cann-bench-eq4-v1}.

\paragraph{Operator completion.}
All 53 operators compile in both case sets. CAKE passes all 1,060 standard
cases ($100\%$) and 4,167 of 4,240 hidden cases ($98.28\%$). All 53 operators
pass every standard case, and 38 pass every hidden case.

\paragraph{Correctness and performance.}
The benchmark score includes performance as well as correctness. CAKE achieves
mean operator scores of 79.31 on the standard set and 77.95 on the hidden set.
The overall mean is 78.63 out of 100, averaging equally over the 53 operators
and the two case sets. Correctness is the property that CAKE's acceptance
criteria enforce; the remaining score reflects opportunities for further
correctness and performance improvement.

\begin{figure}[ht]
  \centering
\begin{tikzpicture}
\begin{groupplot}[group style={group size=2 by 1,horizontal sep=1.5cm},
width=0.46\linewidth,height=5.7cm,ybar=8pt,/pgf/bar width=10pt,
ymin=0,ymax=112,ytick={0,20,40,60,80,100},
xtick={1,2,3,4},xticklabels={L1,L2,L3,L4},
enlarge x limits=0.20,ymajorgrids=true,grid style={gray!20},
axis lines*=left,tick align=outside,/tikz/font=\small,
xlabel={Difficulty level},legend columns=2,
legend style={draw=none,font=\small},
nodes near coords,nodes near coords style={font=\scriptsize},
every node near coord/.append style={/pgf/number format/fixed,
/pgf/number format/precision=1}]
\nextgroupplot[title={(a) Mean operator score},ylabel={Score (out of 100)},legend to name=cannbenchlegend]
\addplot[fill=anaC,draw=none] coordinates {(1,86.85897371) (2,82.07413163) (3,76.31681619) (4,74.11025131)};
\addplot[fill=gray!60,draw=none] coordinates {(1,91.76393501) (2,80.31383741) (3,76.46415560) (4,63.33697710)};
\legend{Standard,Hidden}
\nextgroupplot[title={(b) Case pass rate},ylabel={Passed cases (\%)}]
\addplot[fill=anaC,draw=none] coordinates {(1,100.00000000) (2,100.00000000) (3,100.00000000) (4,100.00000000)};
\addplot[fill=gray!60,draw=none] coordinates {(1,100.00000000) (2,99.21875000) (3,98.21428571) (4,94.84375000)};
\end{groupplot}
\end{tikzpicture}
\par\smallskip\ref{cannbenchlegend}
  \caption{CAKE's aggregated results by difficulty level on Ascend 950PR.
  Levels 1--4 contain 8, 16, 21, and 8 operators. Left: mean operator score,
  with a maximum of 100. Right: case pass rate, computed over 20 standard
  or 80 hidden cases per operator. The same operator grouping is used for
  both sets.}
  \label{fig:cannbench-950pr}
\end{figure}

\paragraph{Score profiles across difficulty.}
Figure~\ref{fig:cannbench-950pr} separates the results by difficulty. Standard-set
case pass rate is $100\%$ at every level, while mean scores are 86.86, 82.07,
76.32, and 74.11 for levels 1--4. On the hidden set, case pass rates are
$100\%$, $99.22\%$, $98.21\%$, and $94.84\%$, with mean scores of 91.76,
80.31, 76.46, and 63.34. Thus, functional coverage remains high across levels,
while the lower scores on the harder tasks identify opportunities for further
kernel optimization.

\subsection{Router Observations}
\label{subsec:router-evidence}

The evaluation ledger illustrates why routing uses operator features as well as
historical scores. In the current ledger,
DSL-A's raw mean exceeds DSL-C's by $0.105$, while the difficulty-adjusted scalar
abilities reported in Section~\ref{subsec:dsl-routing} are statistically
indistinguishable. The per-dimension model then restores operator dependence:
the four worked decisions in Table~\ref{tab:router-decisions} select three
different tracks, with data movement, blocking, and scheduling/reduction
features changing the ordering.

The decision records contain the operator features, feasible tracks, ability
estimates, and skill versions needed to reconstruct each choice. The examples
show how CAKE adapts track selection to the requested operator, rather than
always choosing the track with the highest raw mean score.

\subsection{Operational Evidence for CAKE Book}
\label{subsec:cake-book-evidence}

CAKE CI records the artifacts from which a knowledge entry is created,
CAKE Book preserves its provenance and review state, and skills can retrieve the
result as task-time context. On the write side of this loop, the diagnostics a
campaign produces while debugging one operator are retained in the shared layer
instead of remaining in the case directory, so a later run on an unrelated
operator can retrieve them.

The evaluated CAKE workflow uses CAKE Book together with static references and the
Router. The records complement those end-to-end results by showing how
knowledge is captured, reviewed, and made available to subsequent tasks.

\section{Scope and Future Directions}
\label{sec:scope-roadmap}

The CANN-Bench results cover 53 operators on both standard and hidden case
sets. Router decision records illustrate operator-aware
track selection, the EngramGateFusion campaign demonstrates iterative kernel
optimization, and the CAKE Book records describe cross-task knowledge reuse.
The aggregate benchmark results use Ascend 950PR; the Router examples and
EngramGateFusion campaign retain their Ascend 910B/910C settings.
Appendix~\ref{app:a5} documents the programming paths and generation logic
for Ascend 950.

Further development will extend Evolution to additional operator families and
expand the reusable knowledge available through CAKE Book. Future evaluations
will examine routing and retrieval configurations under matched task sets and
run budgets, and follow knowledge from CI diagnostics through reuse and
reviewed reference updates. The aim is to improve operator coverage and
performance while keeping generation cost and expert review effort practical.

\section{Conclusion}
\label{sec:conclusion}

CAKE organizes Ascend operator engineering around a unit larger than a generated
kernel snippet: a deployable custom operator accompanied by its contract,
reference, lowering artifacts, package, binding, evaluation record, and profile.
Adaptive routing selects among changing code-generation tracks from accumulated
operator evidence; Evolution mode searches within one operator contract using
on-device feedback; and CAKE Book with the Self-Evolving Agent carries lessons
across runs under human review.

The aggregated CANN-Bench results show full standard-set case coverage and
$98.28\%$ hidden-set case coverage across 53 operators on Ascend 950PR.
The EngramGateFusion case study shows that Evolution mode can improve an
already-correct operator under on-device feedback.

\section*{Acknowledgments} We thank Jiwei Yang and Jiong Jin for helpful discussions; Tongtong Xu, Yu Cheng, Yiyu Zhang, and Zhongzhen Wen for insightful discussions on domain-specific languages for NPU kernels; and Chengdi Ma for valuable discussions on SIMT programming and vector functions (VFs). We are grateful to Fengming Zhu for assistance in setting up the continuous integration (CI) pipeline, and to Li Bao, Chengqiu Hu, and Qichen Liao for valuable discussions and suggestions on the generation and optimization of complex operators, including FlashAttention.\par\bibliographystyle{unsrt}
\bibliography{references}

\appendix

\ascendninefiftyappendix

\end{CJK*}
\end{document}